\documentclass[lettersize,journal]{IEEEtran}
\usepackage{amsmath,amssymb,amsfonts}
\usepackage{algorithmic}
\usepackage{graphicx}
\usepackage{textcomp}
\usepackage{algorithm}
\usepackage{array}
\usepackage[caption=false,font=normalsize,labelfont=sf,textfont=sf]{subfig}
\usepackage{textcomp}
\usepackage{stfloats}
\usepackage{url}
\usepackage{verbatim}
\usepackage{graphicx}
\def\BibTeX{{\rm B\kern-.05em{\sc i\kern-.025em b}\kern-.08em
		T\kern-.1667em\lower.7ex\hbox{E}\kern-.125emX}}
\usepackage{balance}

\usepackage{longtable}
\newcolumntype{P}[1]{>{\centering\arraybackslash}p{#1}}

\def\BibTeX{{\rm B\kern-.05em{\sc i\kern-.025em b}\kern-.08em
    T\kern-.1667em\lower.7ex\hbox{E}\kern-.125emX}}
\begin{document}

\title{Automated Semiconductor Defect Inspection in Scanning Electron Microscope Images:\\ a Systematic Review}

\author{Thibault Lechien, Enrique Dehaerne~\IEEEmembership{Graduate Student Member,~IEEE}, Bappaditya Dey~\IEEEmembership{Member,~IEEE}, Victor Blanco, Sandip Halder, Stefan De Gendt~\IEEEmembership{Senior Member,~IEEE} and Wannes Meert~\IEEEmembership{Member,~IEEE}
	
\thanks{Thibault Lechien and Wannes Meert are with the Department of Computer Science, KU Leuven, 3001 Leuven, Belgium.}
\thanks{Thibault Lechien, Enrique Dehaerne, Bappaditya Dey, Victor Blanco, Sandip Halder and Stefan De Gendt are with the Interuniversity Microelectronics Centre (imec),  3001 Leuven, Belgium.}
\thanks{Enrique Dehaerne is with the Faculty of Science, KU Leuven, 3001 Leuven, Belgium.}
\thanks{Stefan De Gendt is with the Department of Chemistry, KU Leuven, 3001 Leuven, Belgium.}
\thanks{Corresponding author: Bappaditya Dey (e-mail: bappaditya.dey@imec.be).}
\thanks{Manuscript received XX/XX; revised XX/XX.}}

\markboth{Preprint}%
{Shell \MakeLowercase{\textit{Lechien et al.}}: Automated Semiconductor Defect Inspection in Scanning Electron Microscope Images: a Systematic Review}

\maketitle

\begin{abstract}
	A growing need exists for efficient and accurate methods for detecting defects in semiconductor materials and devices. These defects can have a detrimental impact on the efficiency of the manufacturing process, because they cause critical failures and wafer-yield limitations. As nodes and patterns get smaller, 
	even high-resolution imaging techniques such as Scanning Electron Microscopy (SEM) produce noisy images due to operating close to sensitivity levels and due to varying physical properties of different underlayers or resist materials.
	This inherent noise is one of the main challenges for defect inspection.
	One promising approach is the use of machine learning algorithms, which can be trained to accurately classify and locate defects in semiconductor samples. Recently, convolutional neural networks have proved to be particularly useful in this regard. 
	This systematic review provides a comprehensive overview of the state of automated semiconductor defect inspection on SEM images, including the most recent innovations and developments.
	38 publications were selected on this topic, indexed in IEEE Xplore and SPIE databases. For each of these, the application, methodology, dataset, results, limitations and future work were summarized. 
	A comprehensive overview and analysis of their methods is provided. 
	Finally, promising avenues for future work in the field of SEM-based defect inspection are suggested.
\end{abstract}


\section{Introduction}
\label{sec:introduction}
\IEEEPARstart{T}{he} manufacturing process of semiconductors is highly complex and involves a variety of different steps, each of which can introduce stochastic defects. 
These can affect electrical performance and reliability, so identifying and addressing these defects at early stages is crucial. 
Furthermore, technology nodes are consistently getting smaller as they continue to follow Moore's law \cite{moore_cramming_1965}.
This reduces the viability of optical inspection methods for detecting small defects. Namely, the resolution of optical methods is not high enough to reliably detect sub 10nm defects \cite{dey_deep_2022-3}.
E-beam inspection methods such as Scanning Electron Microscopy (SEM) have therefore become popular for this purpose. SEM is a high-resolution imaging technique which enables detailed analysis of a sample's surface at the nanoscale. While its throughput is slower than optical methods, its capabilities in imaging make it an invaluable tool for semiconductor defect inspection.

However, detecting defects with SEM is also becoming more challenging. 
First, the extremely small node sizes make it difficult to capture fine details due to the inherent noise involved. 
Even small defects can affect the electrical properties of semiconductor materials, so accurate and robust defect metrology is important \cite{dey_deep_2022-2}.
\begin{figure}[]
	\centering
	\subfloat[]{\includegraphics[width=1.65in]{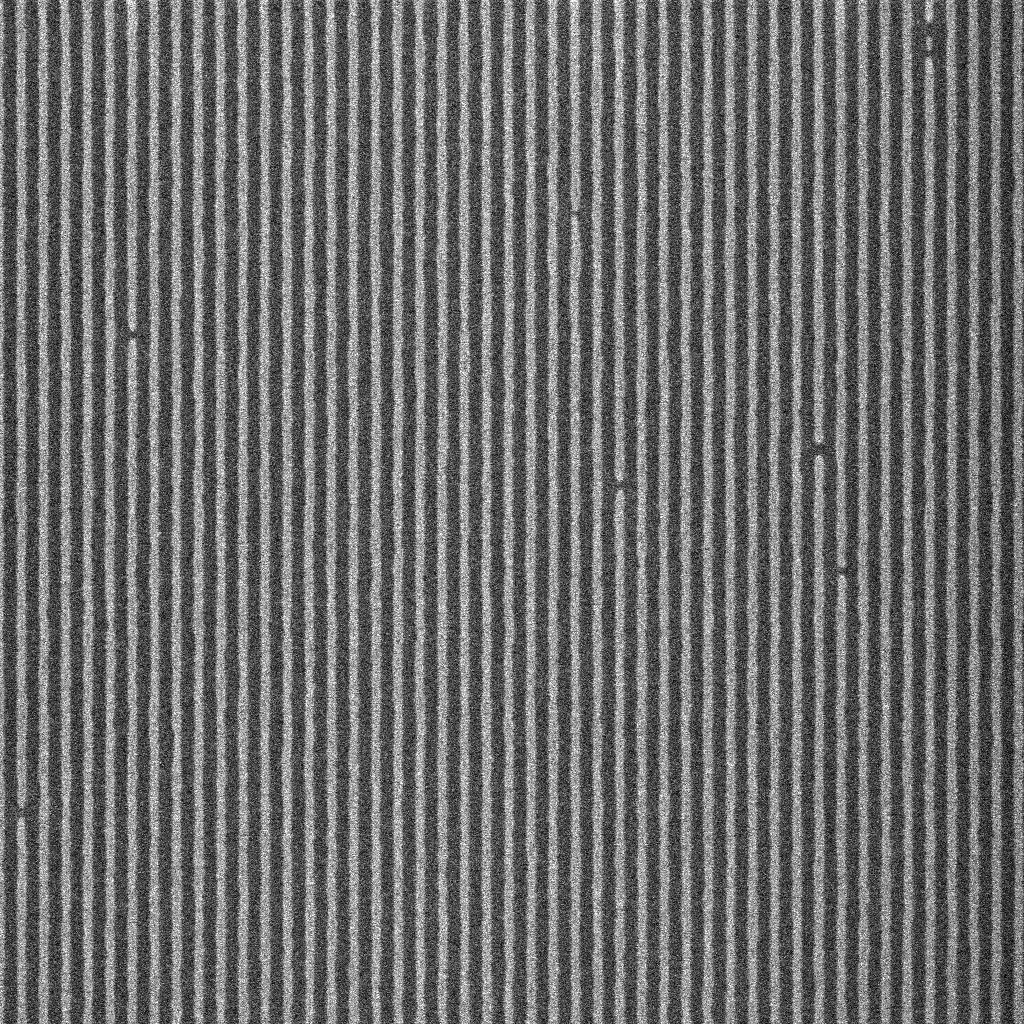}%
		\label{fig: noise1}}
	\hfil
	\subfloat[]{\includegraphics[width=1.65in]{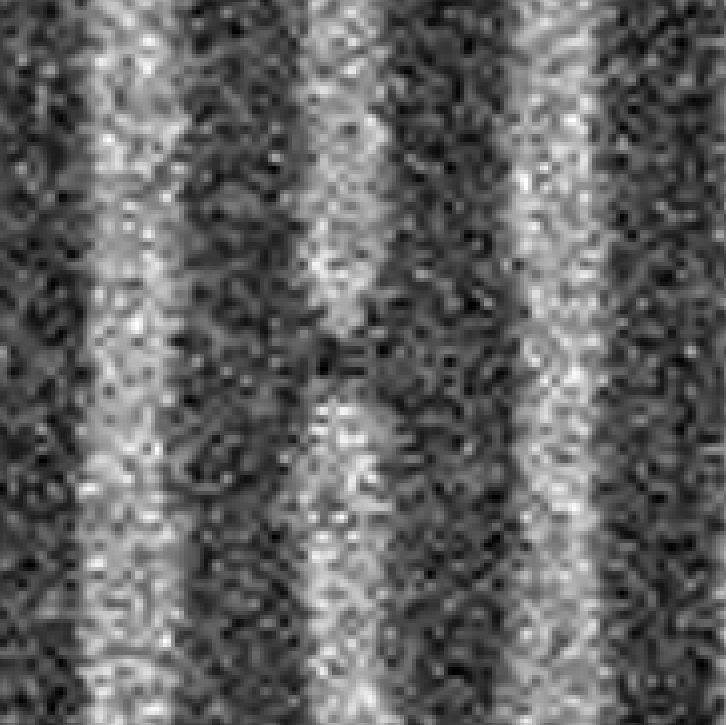}%
		\label{fig: noise2}}
	
	\subfloat[]{\includegraphics[width=1.65in]{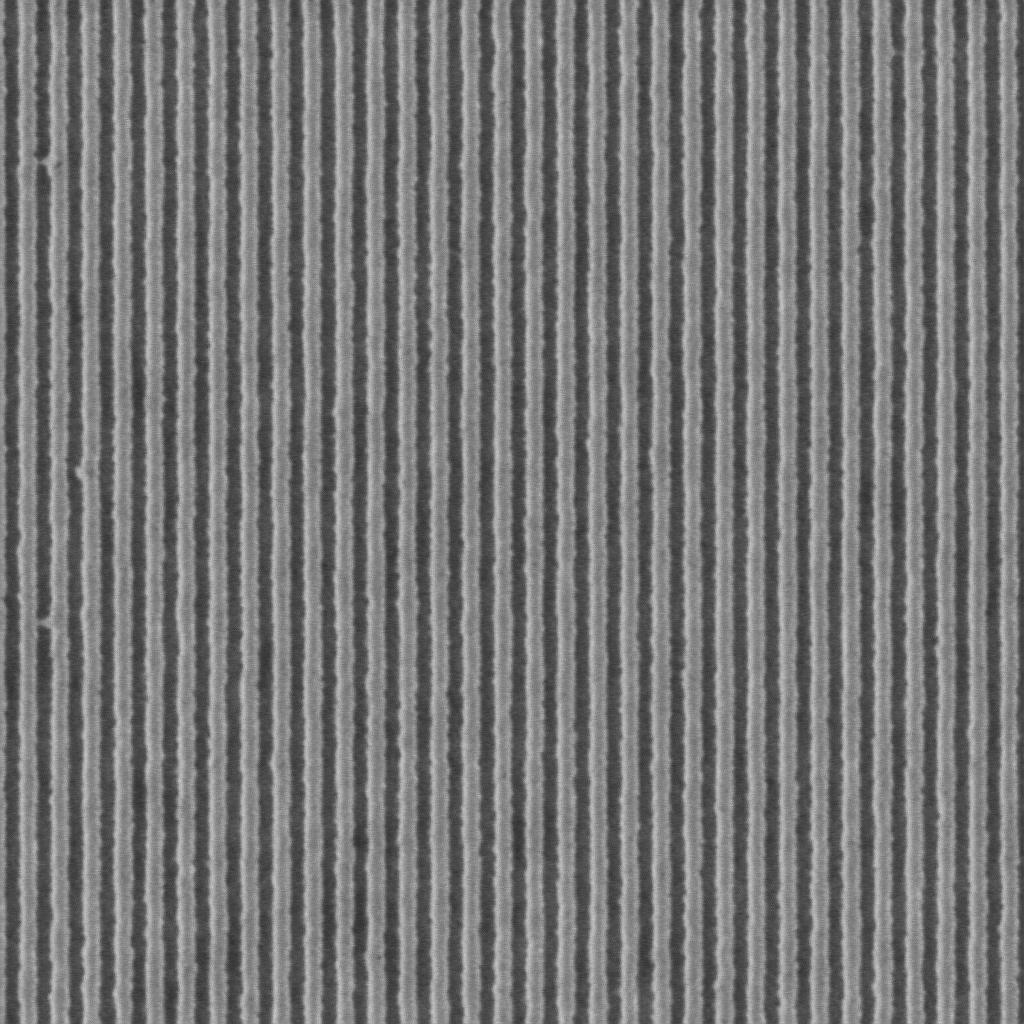}%
		\label{fig: noise3}}
	\hfil
	\subfloat[]{\includegraphics[width=1.65in]{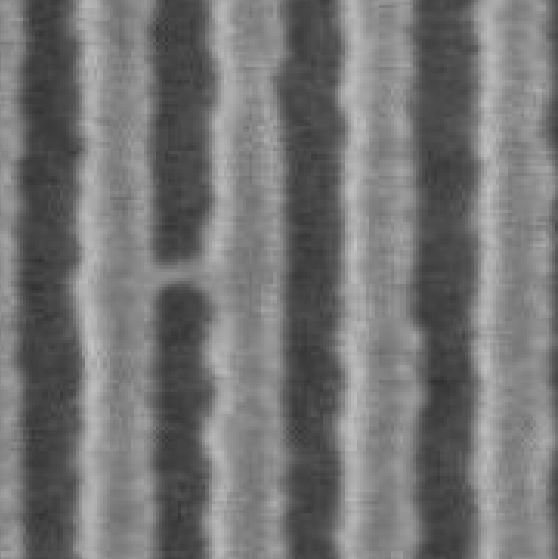}%
		\label{fig: noise4}}
	%\subfloat[]{\includegraphics[width=2in]{Figures/Classification.pdf}%
		%\label{fig: detection_classification}}
	\caption{An illustration of two challenging defect inspection scenarios on line space patterns. (a) A noisy image with gap defects. (b) A zoomed-in image of a gap defect. (c) An image with micro bridge defects. (d) A zoomed-in image of a micro bridge defect. Source: \cite{dehaerne_optimizing_2023}.}
	%(d) Defect classification.}
\label{fig: noise}
\end{figure}
Some visual examples of these challenging scenarios are shown in Fig. \ref{fig: noise}.
Second, manual inspection is not viable due to the large amount of operator time required and the fact that human subjectivity or tiredness introduces biases, which negatively affect detection accuracy \cite{breaux_automatic_1995}. 
Third, although automated detection models have been shown to be more efficient and accurate than manual methods, the wide range of possible defects makes it hard for these models to detect and classify each defect type~\cite{imoto_cnn-based_2019}.
Finally, with the continuous development of new semiconductor materials and fabrication techniques, previously unseen defects are emerging, making it more challenging for defect detection methods to keep up~\cite{cheon_convolutional_2019}. This requires constant innovation and improvement of defect inspection techniques to ensure that semiconductor materials are defect-free and meet the increasing demands of the industry.

One potential solution to these challenges is the use of machine learning algorithms that can learn to identify and classify defects in semiconductor materials. By using large amounts of data to train these algorithms, they can become more accurate and efficient at detecting defects, and can even learn to identify previously unknown defect types.
A promising technique is the use of convolutional neural networks (CNNs), which have seen a rise in popularity in recent years~\cite{cheon_convolutional_2019,imoto_cnn-based_2019,fujishiro_minimizing_2021,dey_deep_2022-1,ofir_automatic_2022}. CNNs are a type of machine learning algorithm that are particularly well-suited for image recognition tasks, making them a promising solution for defect inspection.

The primary objective of this systematic review is to present a thorough and up-to-date analysis of the current state of automated semiconductor defect inspection in SEM images. 
This is done by providing a comprehensive overview of the various techniques and methodologies employed in this field. It also highlights the challenges and limitations of existing approaches, and explores the potential of emerging technologies and techniques, such as machine learning algorithms.
This review also aims to guide future research efforts by identifying areas where further work is needed, which can help to drive innovation and progress in the field of automated semiconductor defect inspection.

There are a few related systematic reviews that focus on different aspects of defect inspection in the semiconductor domain.
Examples include a review of deep learning methods for silicon wafer defect recognition \cite{batool_systematic_2021}, a review of machine learning based methods for defect detection in wafer maps \cite{theodosiou_review_2023}, and a review on wafer surface defect detection methods \cite{ma_review_2023}, but they do not focus on SEM images.
There is also a review of deep learning techniques for surface defect inspection of industrial products~\cite{zheng_recent_2021}, but it is not focused on the semiconductor domain. Instead, it encompasses a broader range of industrial products including steel and fabric.
Lastly, there is a closely related systematic review on machine learning methods for SEM image defect detection and classification \cite{lopez_de_la_rosa_review_2021}. However, it only included 9 studies published until 2020, so it was not able to cover the large amount of recent machine learning work.
In contrast, this review covers 38 studies and goes beyond machine learning methods that perform both detection and classification simultaneously, thus providing a more comprehensive view of the field. 

The remainder of this paper is structured as follows: first, the methodology used to select publications for this review is given in Sec.~\ref{S: methodology}. Next, the results and a summary for each included publication are shown in Sec.~\ref{S: results} and thoroughly discussed in Sec.~\ref{S: discussion}, where an overview is given on the main applications, methods and trends. Finally, directions for future work are suggested before concluding this systematic review.

\section{Methodology}\label{S: methodology}
This review follows the Preferred Reporting Elements for Systematic Reviews and Meta-analyses (PRISMA) guidelines \cite{page_prisma_2021}. The IEEE Xplore \cite{noauthor_ieee_nodate} and SPIE Digital Library \cite{noauthor_spie_nodate} databases were searched using specific inclusion and exclusion criteria. 
Each database was searched on the $19^{\text{th}}$ of April 2023 using the following search terms (each line connected using the `AND' operator): 
\begin{itemize}[]
	\item `defect' OR `flaw' OR `fault' OR `anomaly' OR `deficiency' OR `impurity' OR `contaminant',
	\item `inspection' OR `analysis' OR `detection' OR `localization' OR `segmentation' OR `identification' OR `classification',
	\item `semiconductor' OR `wafer' OR `circuit',
	\item `SEM' OR `scanning electron microscope' OR `scanning electron microscopy'.
\end{itemize}
Note that the term `automated' was not included because some publications assume this is implicit in modern-day defect inspection.  An additional inclusion criterion is that only conference proceedings or journal articles were considered, e.g. no posters, patents or e-books.
After identifying potential publications, they were each examined according to three exclusion criteria. These criteria are as follows: 
%(1) papers that do not examine an automated inspection method, %not examined or not automated
%(2) papers that use other data in addition to SEM images (e.g. metadata or a CAD layout), except if this data is not strictly necessary for the proposed method, and
%(3) papers that do not use SEM images (or only for post-detection evaluation).
\begin{enumerate}[]
	\item papers that do not examine an automated inspection method,
	\item papers that use other data in addition to SEM images (e.g. metadata or a CAD layout), except if this data is not strictly necessary for the proposed method, and
	\item papers that do not use SEM images (or only for post-detection evaluation). 
\end{enumerate}
Publications that met any of these criteria were excluded. First during an initial screening of just their title and abstract, and second during a final screening of the full text. If the title and abstract did not provide sufficient information to justify a publication's removal during the initial screening, it was kept for further examination in the final screening.

\section{Results}\label{S: results}
\begin{figure} [H]
	\centering
	\includegraphics[width=.99\columnwidth]{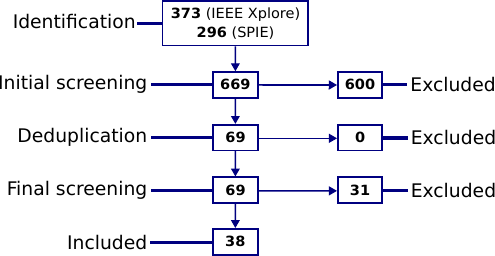}
	\caption{Flow diagram which shows the number of publications at each step of the identification, screening, deduplication and inclusion process.}
	\label{fig:flow_diagram}
\end{figure}

The identification of potential publications using the search methodology detailed in the previous section resulted in 373 and 296 publications from IEEE Xplore \cite{noauthor_ieee_nodate} and the SPIE Digital Library \cite{noauthor_spie_nodate} respectively. The initial screening of their titles and abstracts according to the exclusion criteria resulted in 69 publications (of which 47 are from IEEE Xplore and 22 are from SPIE). No duplicates were found in these publications. A final screening of the full texts resulted in 38 selected publications (of which 19 are from IEEE Xplore and 19 are from SPIE). A flow diagram that shows the number of publications at each step of the selection process can be seen in Fig.~\ref{fig:flow_diagram}.

The selected publications were partitioned into three broad categories according to their methodology.
The first are non-machine learning (non-ML) based methods, second are  machine learning (ML) based methods (excluding convolutional neural networks) and third are convolutional neural network (CNN) based methods.

A summary of each publication in these categories can be found in Tables \ref{tab:nonML}, \ref{tab:ML} and \ref{tab:CNN}, where their application, methodology, dataset, results, limitations and future work are outlined. Publications (within a table) are chronologically ordered by publication date and their Crossref \cite{noauthor_crossref_nodate} citation count is given. 
Note that a direct comparison of their results is not possible because each publication has a different application, uses a different dataset, or uses different evaluation metrics. The next section contains a more general discussion.

Common abbreviations used in Tables \ref{tab:nonML}, \ref{tab:ML} and \ref{tab:CNN} are as follows (in alphabetical order): 
\textbf{ADC} (Automated Defect Classification), 
\textbf{AP} (Average Precision),
\textbf{CD-SEM} (Critical Dimension Scanning Electron Microscope),
\textbf{CNN} (Convolutional Neural Network),
\textbf{EUV} (Extreme Ultraviolet),
\textbf{GAN} (Generative Adversarial Network),
\textbf{kNN} (k-nearest neighbors), 
\textbf{ML} (Machine Learning),
\textbf{PNN} (Probabilistic Neural Network),
\textbf{R-CNN} (Region-based Convolutional Neural Network),
\textbf{SEM} (Scanning Electron Microscope),
\textbf{SMOTE} (Synthetic Minority Oversampling Technique) and
\textbf{SVM} (Support Vector Machine).

\section{Discussion}\label{S: discussion}
This section discusses key findings, trends and challenges of the 38 selected studies. First, the different applications are introduced in Sec. \ref{SS: application} and common metrics in Sec. \ref{SS: metrics}. Next, each method category is thoroughly examined in Sec. \ref{SS: methods}, \ref{SS: nonML}, \ref{SS: ML}, and \ref{SS: CNN} where their main benefits and limitations are also given. General trends are identified in Sec. \ref{SS: trends} and suggestions for future work are given in Sec. \ref{SS: future}.

\subsection{Application}\label{SS: application}
There are two main applications in the selected studies, namely defect detection and defect classification.

\begin{figure}[]
	\centering
	\subfloat[]{\includegraphics[width=2in]{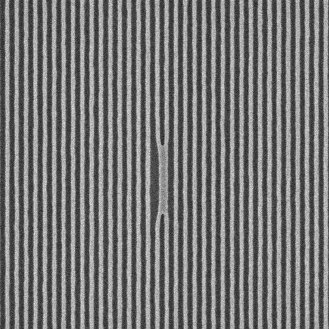}%
		\label{fig: detection_input}}
	\hfil
	\subfloat[]{\includegraphics[width=1.65in]{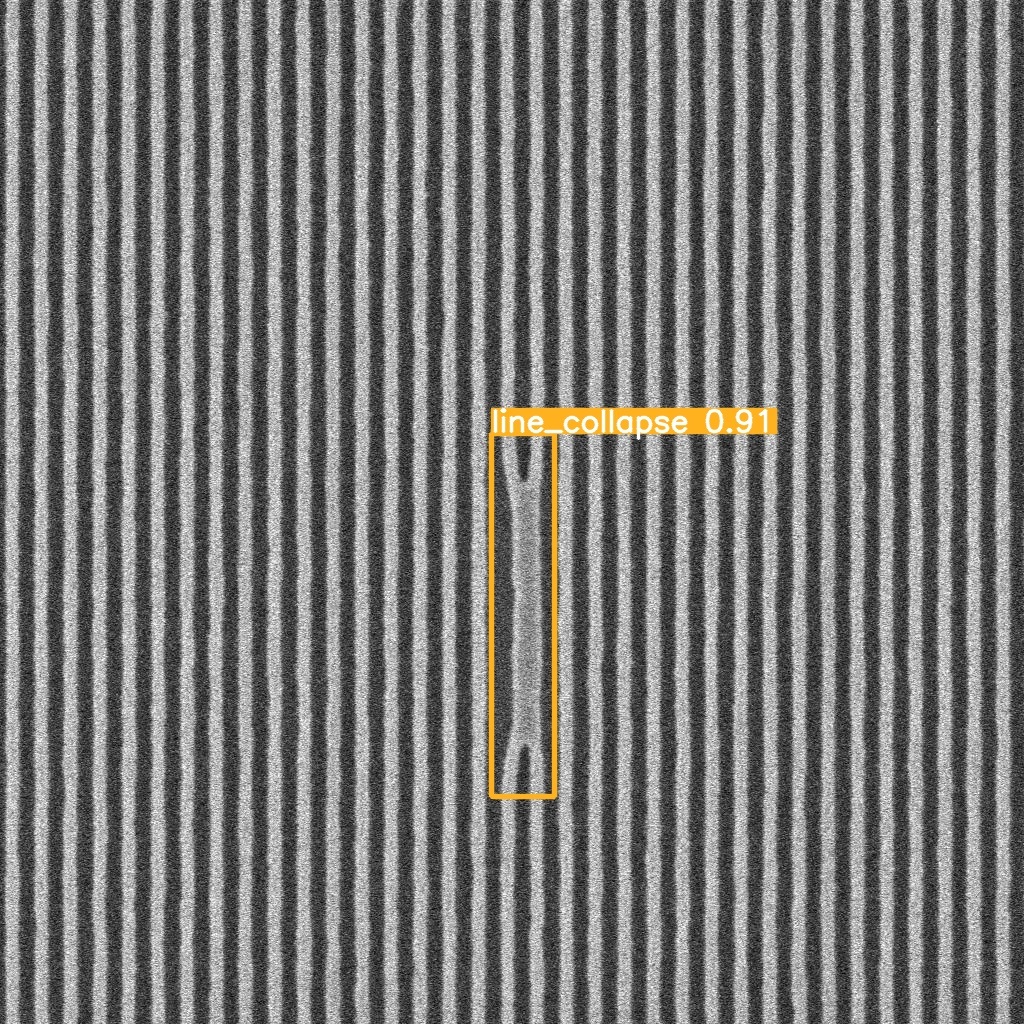}%
		\label{fig: detection_detection}}
	\hfil
	\subfloat[]{\includegraphics[width=1.65in]{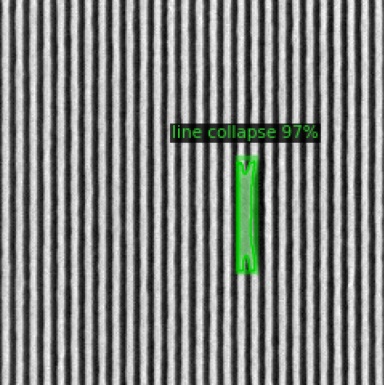}%
		\label{fig: detection_segmentation}}
	\hfil
	%\subfloat[]{\includegraphics[width=2in]{Figures/Classification.pdf}%
		%\label{fig: detection_classification}}
	\caption{An illustration of defect inspection, where a defective input image of a line-space pattern with a line collapse defect type is given as an example. (a) An input image. (b) Defect detection and classification. 
		(c) Defect segmentation and classification.}
	%(d) Defect classification.}
\label{fig: detection}
\end{figure}

Defect detection aims to determine whether there is a defect in a given image and where it is located. 
Defect classification, on the other hand, aims to identify the type of defect present in the image. 
Note that some publications defined defect detection as a binary classification problem, where the two classes are `defective' and `non-defective', but this definition is not used in this review. Instead, defect detection is defined based on bounding boxes, which is similar to object detection in the broader domain of computer vision. There are also some other applications such as defect segmentation, which aims to identify defects on a pixel-level, but they are somewhat uncommon in the included studies and grouped under the `Other' application type. 
An illustration of common application types is given in Fig. \ref{fig: detection}, where a defective input image and an ideal output is given for  defect detection, classification and segmentation tasks.

\begin{figure} []
\centering
\includegraphics[width=.9\columnwidth]{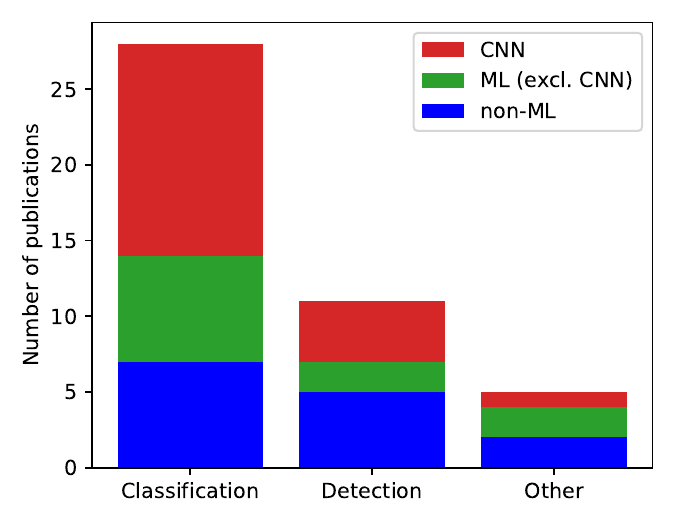}
\caption{A histogram of the number of publications in each of the three categories in terms of their main application. If a publication has multiple applications, it is counted in each of them.}
\label{fig:studies over application}
\end{figure}

A histogram of the number of publications in terms of their main application and method category can be seen in Fig.~\ref{fig:studies over application}. If a study has multiple main applications, they are counted in each application. Classification is the most common application, likely because knowing the defect type is useful for analyzing its source.
Detection is less common and segmentation even less again, which is likely due to the added difficulty of localizing defects on a pixel-level. Within each application, the proportions of each method category (non-ML, ML and CNN) are roughly equal, indicating that each method can be used for both of the main applications.
These methods will be examined in depth in the following sections.

%% Important: tables should fit on exactly an integer amount of pages
% If not possible: use same bigtable as dehaerne et al.
\onecolumn
\footnotesize
\begin{longtable}{ P{0.7cm}|p{2.5cm}|p{3.5cm}|p{3cm}|p{3cm}|p{3cm}}
	\caption{Non-machine learning based automated defect inspection}\\
	Study & Application & Methodology & Dataset & Results \& Discussion & Limitations \& Future Work \\
	\endfirsthead
	\caption*{Non machine learning based automated defect inspection (continued)}\\
	Study & Application & Methodology & Dataset & Results \& Discussion & Limitations \& Future Work \\
	\endhead
	\hline
	\cite{breaux_automatic_1995} (Cited by 14)
	& Defect classification.
	& A simple architecture consisting of feature extraction and (fuzzy) rule application. A defect description language allows engineers to specify defect classes.
	& Wafer images from four different layers (active, gate, metal 1 and metal 2). The number of images was not specified.
	& The proposed method achieved accuracies between 70\% and 95\% depending on the layer and beat humans in terms of repeatability.
	& The slow inference speed is a large limitation. Improving this is suggested as future work. \\
	\hline	
	\cite{ben-porath_advanced_1999} (Cited by 5)
	& Automating in-line defect detection and classification.
	& The SEMVision™ tool.
	& Images of three different layers with 8 to 12 defect classes per layer. The number of images was not specified.
	& The tool achieved an accuracy between 75.3\% and 97.2\% on the three considered layers.
	& Work is being done to generalize this process to unpatterned wafers. \\
	\hline
	\cite{ritchison_sem-based_2000} (Cited by 6)
	& Integrating defect classification in an advanced process fab.
	& The SEMVision™ cX tool.
	& 4 datasets stemming from different inspection levels with 471 to 885 images and 6 to 7 defect types each.
	& The tool achieved an average accuracy of roughly 80\%, ranging from 17\% to 100\% per defect type.
	& Future work is required on tool matching, recipe management issues and improving the user interface. \\
	\hline
	\cite{gleason_detection_2002} (Cited by 6)
	& Defect detection.
	& Fractal image encoding and quadtree partition were used for defect detection. Active contours were applied for better defect boundary delineation.
	& 2077 images from 3 different layers.
	& The proposed method achieved accuracies between 57\% and 92\%  depending on the layer.
	& Small defects in complex backgrounds are often missed and the auto-parameter generation procedure requires further validation. \\
	\hline
	\cite{mcgarvey_automated_2009} (Cited by 0)
	& Automated defect classification of the wafer bevel.
	& Using the Hitachi defect review SEM RS-5500EQEQ's automated defect classification system.
	& 55 defects in 4 sets of 300mm wafers, partitioned into 2 defect types.
	& The proposed method achieved accuracies ranging from 80.6\% to 98.4\% and throughputs that were between 330 to 763 defects per hour.
	& Deciphering defect sources and formulating an approach to reduce the root cause of the defects. \\
	\hline
	\cite{zontak_kernel-based_2009} (Cited by 3)
	& Defect detection.
	& Exploiting the periodic nature of wafers with a kernel-based detection procedure, without the need for reference images.
	& Multichannel SEM images. The number of images was not specified.
	& Promising results were found compared to other widely used methods.
	& Runtime was a large limitation, so future work could include optimizing the method. Concrete accuracy numbers are also desirable.\\
	\hline
	\cite{lee_accurate_2014} (Cited by 6)
	& Defect detection.
	& 5 steps consisting of alignment, edge relaxation, quadtree decomposition, multi-channel integration and defect scoring.
	& 1000 images containing 4 different defect types and 600 defect-free images.  %(512x512)
	& More reliable and robust results were achieved than conventional methods, lowering false positive and negative rates from 4.3\% to 3.7\% and from 7.0\% to 2.3\% respectively.
	& Inference time was a limitation, so they are planning to parallelize their architecture. Extending it to RGB images would also be interesting. \\
	\hline
	\cite{samir_improvement_2015} (Cited by 1)
	& Defect classification and size estimation.
	& Calibre®'s MDPAutoClassify™ tool that extracts defect signals, combines them into properties and uses them in classifiers.
	& CD-SEM images of programmable defect and production masks. The number of images was not specified.
	& The method achieved an overall classification accuracy of 97\% and an accuracy of 99\% for critical defects.
	& Future work includes assessing and improving the tool's applicability to EUV masks. \\
	\hline
	\cite{cho_classification_2017} (Cited by 1)
	& EUV mask defect classification and printability assessment.
	& KLA-Tencor’s Reticle Decision Center which calculates the difference between a reference and a (denoised) test image and extracts defect signatures.
	& 181 images with programmed defects and 1328 with natural defects.
	& An accuracy of 99.5\% and 95.8\% was achieved on the programmed and natural defects respectively.
	& A future step is to correlate print accuracy with actual wafer prints. \\
	\hline
	\cite{wolfe_scanning_2017} (Cited by 0)
	& Process-induced defect classification.
	& Using the Hitachi defect review SEM RS4000's automated defect classification system.
	& Images from multiple wafers from 4 different film stacks. The number of images was not specified.
	& An accuracy of 91.3\% to 99.2\% was achieved across the different film stacks.
	& Low contrast images were challenging to classify, so further work could focus on improving this scenario. \\
	\hline
	\cite{egodage_sem_2019} (Cited by 0)
	& Defect detection.
	& SEM AutoAnalysis which consists of 3 steps: contour extraction, image correlation with a reference image, and defect identification.
	& 12 images with a lines and spaces pattern and 6 images with a contact hole pattern. %(1898×1489)
	& "Better-than-human" performance was achieved in terms of inference time and reliability.
	& Applying this method to other tools and data types can further extend the degree of automation, which is an avenue of further work. \\
	\hline
	\label{tab:nonML}
\end{longtable}
%\twocolumn

\newpage

\begin{longtable}{ P{0.7cm}|p{2.5cm}|p{3.5cm}|p{3cm}|p{3cm}|p{3cm}}
	\caption{Machine learning based automated defect inspection (excluding CNNs)}\\
	Study & Application & Methodology & Dataset & Results \& Discussion & Limitations \& Future Work \\
	\endfirsthead
	\caption*{Machine learning based automated defect inspection (excluding CNNs) (continued)}\\
	Study & Application & Methodology & Dataset & Results \& Discussion & Limitations \& Future Work \\
	\endhead
	\hline
	\cite{hunt_paradigm_2000} (Cited by 0)
	& Defect classification.
	& kNN, PNN and decision trees (separately).
	& 5 datasets containing between 69 and 516 defects each, split into 5 to 8 defect types.
	& The kNN and PNN methods were able to provide complementary performance to the decision trees, so a smart ensemble method might be possible.
	& Future work will investigate methods to automatically transition from a rule-based to a PNN/kNN approach. \\
	\hline
	\cite{hunt_optimizing_2000} (Cited by 2)
	& Defect classification.
	& kNN, fuzzy kNN and a radial basis function-based classifier.
	& 664 images of defects, partitioned into 7 defect types.
	& A highest overall accuracy of 83.54\% was achieved by kNN, showing that non-parametric classifiers can be viable.
	& The considered pair-wise classifier was not found to perform better than kNN, but can be improved in future work. \\
	\hline
	\cite{chang_unsupervised_2005} (Cited by 1)
	& Defect detection.
	& An unsupervised self-organizing neural network \cite{kohonen_self-organizing_2001} that classifies the wafer image into 4 classes and a heuristic to determine which class is defective.
	& Zoomed-in wafer images. The number of images was not specified. %(640x480)
	& The method achieved an average sensitivity and specificity of 97\% and 100\%.
	& Only a limited dataset was used. Additionally, improving the inspection time is desirable. \\
	\hline
	\cite{chang_using_2005} (Cited by 5)
	& Defect segmentation.
	& An unsupervised two-layer competitive Hopfield neural network \cite{hopfield_neural_1982}.
	& 4 types of zoomed-in wafer images. The number of images was not specified.%(640x480)
	& An average sensitivity and specificity of 99.4\% and 98.3\% was achieved.
	& Future work could include extending the limited dataset in order to study the applicability under different scenarios. \\
	\hline
	\cite{chang_similarity_2014} (Cited by 1)
	& Die defect segmentation and classification.
	& A mean absolute deviation\cite{noauthor_mean_2008} method is used to align test and reference images, resulting in a binary difference image. A neural network-based adaptive resonance theory~1~\cite{carpenter_adaptive_2010} method is then used for classification .
	& 16 images containing 104 defects.
	& Reliable and consistent defect classification results were achieved.
	& The dataset used is quite limited and no concrete accuracies were given, so a more thorough analysis is required. \\
	\hline
	\cite{graur_new_2015} (Cited by 2)
	& Detection and classification of yield-impacting EUV mask defects.
	& Local features are extracted through histograms of oriented gradients, upon which 3 sequential binary support vector machines were trained.
	& 1100 images with 3 defect types. Multiple defects per image are possible.
	& An accuracy of 90\% was achieved. The method is easily scalable in terms of defect types and can capture defects that are close to noise level.
	& The performance diminished with few training examples, so increasing learning efficiency can be a next step.\\
	\hline
	\cite{rana_deep_2018} (Cited by 1)
	& Blurred versus non-blurred image classification in hard disk drive manufacturing.
	& Principal component analysis is used for dimensionality reduction, after which a four layer feed-forward neural network classifies the image.
	& 1066 CD-SEM images. %(1000x1600) -> (150x320) using PCA
	& An accuracy of 97.1\% was achieved, outperforming a different image analysis method.
	& More work is required to handle images shifting, for example by using convolutional neural networks. Using Fourier transformations might also be interesting. \\
	\hline
	\cite{lee_automated_2020} (Cited by 1)
	& Defect classification with imbalanced data.
	& The SMOTE oversampling technique \cite{chawla_smote_2002} is used for the training of 4 machine learning algorithms (random forests~\cite{ho_random_1995}, AdaBoost~\cite{freund_desicion-theoretic_1995}, XGBoost~\cite{chen_xgboost_2016} and SVM~\cite{cortes_support-vector_1995}). 
	& 2 datasets: a line pattern set with 1666 images and 10 defect classes, and an unspecified set containing 13737 images and 2 defect classes.
	& A higher average class-wise accuracy was found by using SMOTE, reducing the amount of required training data and lowering the sensitivity of classifier hyperparameter selection.
	& Testing the performance of more advanced machine learning algorithms using SMOTE is proposed as future work. \\
	\hline
	\cite{gomez-sirvent_optimal_2022} (Cited by 4)
	& Optimal feature selection for defect classification.
	& A support vector machine classifier using an optimal feature subset.
	& 5475 images partitioned into 7 defect types.
	& An F1-score of 97.3\% was achieved. They also showed that there is a lot of correlation between features.
	& The feature selection is computationally expensive and restricted to low dimensional datasets. Combining this model with a CNN could yield better results than either model alone and could be an interesting line of research for future work. \\
	\hline
	\label{tab:ML}
\end{longtable}

\newpage
%% Should fit on exactly an integer amount of pages
%\onecolumn
\begin{longtable}{ P{0.7cm}|p{2.5cm}|p{3.5cm}|p{3cm}|p{3.5cm}|p{3cm}}
	\caption{Convolutional neural network based automated defect inspection\label{zzz}}\\
	Study & Application & Methodology & Dataset & Results \& Discussion & Limitations \& Future Work \\
	\endfirsthead
	\caption*{Convolutional neural network based automated defect inspection (continued)}\\
	Study & Application & Methodology & Dataset & Results \& Discussion & Limitations \& Future Work \\
	\endhead
	\hline
	%CNN in chronological order:
	\cite{ducote_microlens_2018} (Cited by 0) %CD-SEM, very little info
	& Microlens under-melt classification.
	& CNN for two-class classification: output is `good' or `bad'.
	& CD-SEM images: defect-free ones are taken from production data, defective ones are generated by undermelting wafers. The number of images was not specified.
	& The method achieved an inference time of less than one minute and is able to stop bad lots in the fab automation system.
	& Increasing reliability: reducing false positives and negatives through prediction review and manual classification. \\
	\hline
	\cite{cheon_convolutional_2019} (Cited by 104)
	& Wafer surface defect classification, including unknown classes
	& CNN with 4 convolutional layers, 2 max-pooling layers, a categorical cross entropy loss function and softmax classification output. Can classify previously unknown defects by using kNN on the CNN training result's feature space. %(160x160)
	& 2153 SEM images sampled from an inspection facility were put into two datasets: one for training and testing the CNN and one for evaluating the detection of unknown classes (not present in the other dataset). Data augmentation was used to triple the training data for the CNN.
	& The average image classification accuracy of the CNN is 96.2\% (each class between 90.7\% and 100\%). The method also achieved 93\% accuracy on the unknown defect classes, which shows that clustering can be viable for ADC.
	& A CNN needs retraining over time to ensure detection of new defects, which is computationally expensive. They propose to examine unsupervised clustering in the future, which will likely be more efficient in this regard. \\
	\hline
	\cite{imoto_cnn-based_2019} (Cited by 34)
	& Defect classification.
	& CNN with 33 convolutional layers (using the Inception architecture \cite{szegedy_rethinking_2015}) and softmax classification output. It is first trained on inaccurate labels and then finetuned on reliable labels.
	& Dataset with inaccurate labels for pretraining (97880 images), part of which was relabeled with reliable labels for finetuning (11988 training images, 11986 testing images). %Images resized to (128x128)
	& The method achieved an average accuracy of 87.26\% and showed that pretraining can improve the classification accuracy, even when the labels are inaccurate. Pretraining was found to be particularly beneficial where the amount of training data was limited.
	& Some classes had poor precision and recall (e.g. between 0 and 60\%), so it should not be relied upon in these cases. Examining possible causes or solutions for this problem are avenues for further work. \\
	\hline
	\cite{dey_unsupervised_2020} (Cited by 1)
	& Image classification for optimal proximity correction and process window estimation.
	& A variational autoencoder with 3 pairs of convolutional and max pooling layers that learned characteristic features and k-nearest neighbors for classification.
	& 230 CD-SEM wafer images. Data augmentation was used.
	& The proposed approach was shown to be effective for deriving the printable window without needing CD-measurements.
	& Improving the model accuracy and performance by fine-tuning different hyper-parameters and to look for stochastic defects during the review of defect locations. \\
	\hline
	\cite{phua_semiconductor_2020} (Cited by 2)
	& Wafer surface defect classification, particularly on the metal layers.
	& ResNet101-based \cite{he_deep_2015} CNN pretrained on ImageNet \cite{deng_imagenet_2009} to utilize transfer-learning and an SSD architecture \cite{liu_ssd_2016} (based on VGG-16 \cite{simonyan_very_2015}) used in a second phase for sizing classification for 2 specific classes. %(400x400) resized to (224x224) for resnet, and to (300x300) for SSD
	& 8943 metal layers defect Review-SEM images from X-FAB containing 8 defect classes,  2 of which were split into 4 subclasses which represent the same defects but with different sizes. Additionally, data augmentation was used.
	& The proposed system achieved a 91.1\% Top-1 accuracy and a 96.2\% Top-3 accuracy, with each class being classified correctly at least 69\% of the time. 
	& A somewhat small, non-uniform dataset was used. Future work will explore using more advanced CNNs on larger, more uniform datasets. \\
	\hline
	\cite{fujishiro_minimizing_2021} (Cited by 0)
	& Inline defect classification with minimal training data, in particular how to use data augmentation effectively.
	& VGG-16 based CNN that was pretrained on ImageNet with 13 convolutional layers, 5 pooling layers, categorical cross entropy as loss function and a softmax classification output.
	& 1432 images were used to construct 6 datasets, each with a different limit on the maximum amount of images in each of the 10 defect classes. Data augmentation was used. %(192x192)
	& An overall accuracy of 94.2\% was achieved, compared to 79.4\% without data augmentation. The greatest benefit of data augmentation was found in the smallest datasets. They showed that reducing labeling costs without sacrificing accuracy is viable.
	& The proposed method is quite specific to the problem, but it can be interesting to study if this can also be applied to other inspection steps. Establishing a general-purpose data augmentation method is desirable. \\
	\hline
	\cite{lei_convolutional_2021} (Cited by 1)
	& Defect classification with imbalanced data.
	& CNN with 3 convolutional layers, 3 max pooling layers and cross entropy as a loss function. 
	& 1091 images partitioned into 8 defect types. A GAN \cite{goodfellow_generative_2014} was trained to generate new images of each type to 1) increase the amount of training data and 2) to get perfectly balanced classes. %(64x64)
	& An accuracy of 95\% was reached by increasing the amount of training data with the GAN, up from 87\% originally. They also showed that perfect class balance is not required.
	& One class dropped by 20\% in the purity measure after using the GAN, which was left undiscussed. It might be interesting to study whether there are any differences between using more `real' training data versus more GAN-generated training data. \\
	\hline
	\cite{lin_improvement_2021} (Cited by 1)
	& Defect classification for multi-lines bridges. 
	& A hierarchical structure of two (binary) ResNet-based CNNs: one to distinguish bridges from other types and then one to distinguish multi-lines bridges from other bridges. 
	& 7200 images were partitioned into 6 defect types. Data augmentation was used in order to increase the data to 45000 images.
	& Compared to a non-hierarchical structure, a hierarchical structure was able to improve the average purity of all classes, especially the multi-lines bridge purity (from 55\% to 88\%). The accuracy was left unchanged. 
	& Training the proposed architecture on more data or applying this method to other processes with similar issues are avenues of further work. \\
	\hline
	%New from ML table:
	\cite{kim_adversarial_2021} (Cited by 4)
	& Defect detection.
	& A CNN-based GAN \cite{mirza_conditional_2014} using a ResNet-18 backbone, focal loss \cite{lin_focal_2018}, a PixelGAN \cite{makhzani_pixelgan_2017} discriminator and multi-scale level features. 
	& 17125 images containing 2 defect types. Data augmentation was used. %(512x512)
	& The proposed method achieved a score of 92.3 in the AP metric. Ablation experiments showed that the chosen loss type, discriminator architecture and features all play key parts in reaching this performance.
	& Testing this approach on more defect types can be interesting. \\
	\hline
	\cite{wang_defect_2021} (Cited by 1)
	& Defect classification.
	& CNN (not specified further).
	& 536 defective images were used for training.
	GAN's (which were trained on 1253 defect patches) were then used to increase the training data to 963 images by turning defect-free images into defective images.
	& An accuracy of 96.2\% was achieved, compared to 90.1\% of the model that trained on the unaugmented data. This approach is also useful for image quality enhancement.
	& A comparison of this approach to other GAN-based data augmentation approaches would be interesting. A more thorough analysis of the change in model output is also desirable. \\
	\hline	
	\cite{fukuda_anomaly_2021} (Cited by 0)
	& Defect detection in random circuit patterns.
	& A convolutional variational autoencoder is applied to a decomposed image. Discrepancies between its input and output are used to detect anomalies. 
	& 675 images from 11 exposure fields. %(1024x1024)
	& Promising preliminary results were found both in terms of accuracy and training/inference time.
	& Testing the proposed method on a large amount of images will be studied as more data becomes available. \\
	\hline
	\cite{dey_deep_2022-1} (Cited by 0)
	& Defect detection and classification.
	& RetinaNet-based detector \cite{lin_focal_2018} using an ensemble of ResNet backbones and a U-Net-based denoiser \cite{ronneberger_u-net_2015}.
	& 1324 images with 5 defect types. Multiple defects are possible in one image. Data augmentation was used. %(1024x1024)
	& Stable performance was achieved for both noisy and denoised images, beating conventional approaches. They reached a value of 0.816 in the mean average precision (mAP) metric.
	& Applying the approach to other manufacturing steps, experimenting with other architectures to further increase accuracy and add new defect categories. \\
	\hline
	\cite{dey_deep_2022} (Cited by 0)
	& Defect detection and classification.
	& Mask R-CNN \cite{he_mask_2018} with a ResNet backbone.
	& 600 images partitioned into 100 images for each of the 6 defect types. Data augmentation was used. %(480x480)
	&  The method achieved a value of 0.936 in the mean average precision (mAP) metric for bounding boxes and segmentation.
	& A perfect class balance scenario was considered, but no experiments with class imbalance were done. The future work is the same as for \cite{dey_deep_2022-1}. \\
	\hline
	\cite{li_automatic_2022} (Cited by 0)
	& Defect classification.
	& Decision tree of deep neural networks consisting of 8 pretrained ResNets.
	& 2975 images partitioned into 12 defect types. Data augmentation was used.
	& The proposed method was shown to be very accurate, more efficient (both during training and inference) compared to standard CNNs and easily (partially) retrainable, without the need for GPU's.
	& More image augmentation methods could be explored. Using more complex neural architectures would also be interesting. \\
	\hline
	\cite{ofir_automatic_2022} (Cited by 0)
	& Defect segmentation.
	& A U-Net-based CNN with multiple heads and dense contrastive learning.
	& 4000 training images with simulated defects implanted in them and 1000 test images with real defects. Data augmentation was used.
	& The unsupervised method achieved an F-score of 0.65, compared to the scores between 0.73 and 0.87 for supervised methods.
	& The accuracy is still considerably lower than supervised methods. Further work is required to study the use of different architectures. \\
	\hline
	\cite{yan_based_2022} (Cited by 0)
	& Defect detection and classification.
	& A pretrained YOLOv2 model \cite{redmon_yolo9000_2016}. The data used for pretraining was not specified.
	& 88 processed CD-SEM images with enhanced contrast and brightness. %(448x448)
	& The method achieved a test accuracy of 100\% and a fast batched inference time of 23 seconds for 188 images.
	& Future work could include training and evaluating on a large dataset. \\
	\hline	
	\cite{dey_towards_2022} (Cited by 0)
	& Defect detection and classification.
	& An ensemble of 5 different pretrained YOLOv5 \cite{jocher_ultralyticsyolov5_2022} architecture variants. The data used for pretraining was not specified.
	& 1324 images with 5 defect types. Multiple defects are possible in one image. Data augmentation was used.
	& The method reached a value of 0.867 in the mean average precision (mAP) metric, beating their previous framework \cite{dey_deep_2022-1}, particularly in the most difficult defect classes.
	& There is further scope for improvement of the overall mAP metric by retraining the architectures with more	training data as well as through hyperparameter tuning. \\
	\hline	
	\cite{neumann_defect_2023} (Cited by 0)
	& Defect detection and classification.
	& A CNN-based autoencoder for detection, using the first three convolutional layers of an AlexNet~\cite{krizhevsky_imagenet_2012}, pretrained on ImageNet~\cite{deng_imagenet_2009}. A similar model is made for classification by appending a sigmoid layer.
	& 91 images containing 2 defect classes. Multiple defects are possible in one image. 
	& The method achieved a capture rate of 97.2\% and a false positive rate of 1.5\% on the noisiest images. This enables high-throughput inspection, as they achieve good detection results even for fast scan speeds.
	& Further improvements are possible by optimizing the flow and by employing more specifically trained modules at the various steps of the workflow. \\
	\hline
	\label{tab:CNN}
\end{longtable}
\twocolumn
\normalsize

\subsection{Metrics}\label{SS: metrics}
In order to quantify the performance of a method, application-specific metrics are used. 
Classification accuracies are quantified by looking at how many instances of each defect type were classified correctly. This is often measured by the accuracy metric (also known as Top-1 accuracy), which is the fraction of correct classifications, but other metrics are also used. 
Examples include precision, which measures how many observations predicted as one class are actually that class; recall, which is equivalent to the true positive rate; F1-score, which is the harmonic mean between precision and recall; and Top-3 accuracy, which is the fraction of correct classifications in any of the three highest probability answers returned by the method. The latter is only applicable if the method is capable of returning probabilities and multiple answers, but this is often the case for the included studies.

\begin{figure} []
\centering
\includegraphics[width=.7\columnwidth]{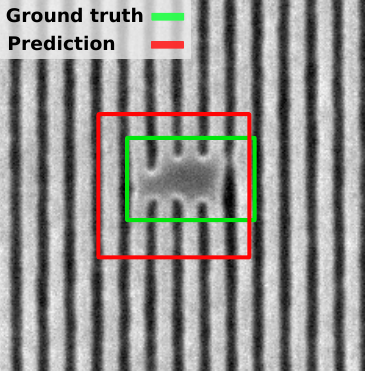}
\caption{An example of a predicted and a true bounding box, where the IOU metric has a value of $0.5$.}
\label{fig:iou5}
\end{figure}

Detection accuracies are quantified by looking at how well the predicted bounding boxes fit the true bounding boxes, which are rectangles that capture the defective regions. This can be evaluated by using the Intersection Over Union (IOU) metric, which calculates the overlap between two bounding boxes. This results in a number between 0 and 1. If the IOU is above a certain threshold, for example 0.5, the prediction is seen as a true positive. An example of an IOU value of $0.5$ is shown in Figure~\ref{fig:iou5}, where the area of the intersection divided by the area of the union of the two bounding boxes is equal to $0.5$.
The IOU is calculated for all predicted defects. However, a model also returns a probability score for each bounding box, which quantifies how confident the model is in that particular prediction.
The Average Precision (AP) metric is then the area under the precision-recall curve, which is obtained by calculating the precision and recall values for different confidence score thresholds \cite{pedregosa_scikit-learn_2011,anwar_average_2022}.
%after which the ratio of true positives over all ground truth defects gives the Average Precision (AP) metric. Alternatively, AP can also be calculated through the area under the precision-recall curve.
If AP is calculated on each defect type separately, the mean Average Precision (mAP) is given by averaging the AP scores of each type. Sometimes AP is calculated for a range of IOU thresholds, which can give more information on how accurate a method is in determining the exact defect size and location.
%TODO: maybe include figure of IOU metric? but is likely evident from the picture already included

\subsection{Methods}\label{SS: methods}
\begin{figure} [h]
	\centering
	\includegraphics[width=.9\columnwidth]{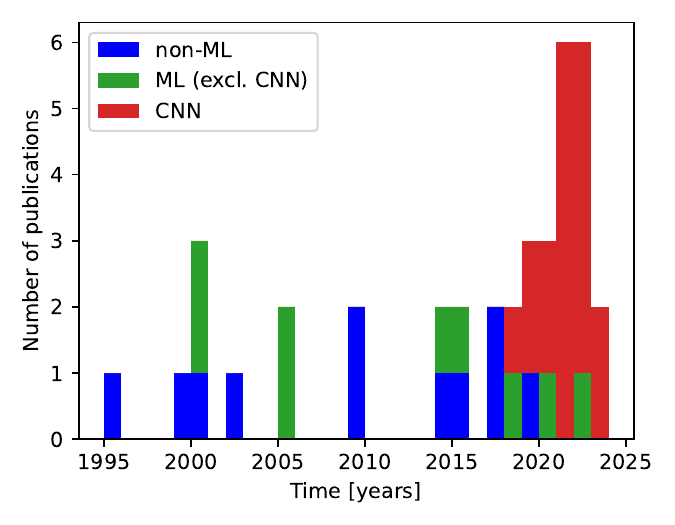}
	\caption{A histogram of the number of publications in each of the three categories in terms of time.}
	\label{fig:studies over time}
\end{figure}

The studies included in this review were partitioned into three categories, namely non-machine learning, machine learning (excluding CNNs), and CNN-based methods. These methodologies have seen varying levels of popularity over the past two decades. This can be seen in Figure~\ref{fig:studies over time} where the number of publications in each category is plotted against time.

The oldest papers included in this review (1995-2000) used manually constructed detection methods, such as manual feature construction and rule application. 
After this, machine learning methods were being introduced as well. During the last few years (2018-2023), the number of studies per year on the topic of defect inspection has increased. This is mainly due to the large number of studies that use convolutional neural networks and its complex architecture variants. These methods will be examined in more detail in following sections.

As mentioned in Sec. \ref{S: methodology}, studies with methods that use other data in addition to SEM images (e.g. metadata or CAD layouts) were excluded, unless the extra data is not strictly required. Instead, some of these studies are briefly discussed here. Examples include \cite{esposito_analysis_2020} and \cite{shah_design_2018}, which proposed a design-based automatic defect classification framework that combines SEM images with CAD information; \cite{oleary_deep_2020} uses energy-dispersive x-ray spectroscopy data to help classify the chemical composition of defects; and \cite{arena_exploiting_2021} uses 2D coordinates of the images as Bayesian priors. Although additional data is not always available or used, these studies show that they can improve detection and classification accuracies.
Note that \cite{kim_adversarial_2021} can use design information, but as the model also works without it, it was included in the review.

\subsection{Non-machine learning based methods}\label{SS: nonML}
The first main method category is non-machine learning based methods, given in Table \ref{tab:nonML}. These are static approaches, meaning that they do not learn from data. 
Instead, they typically rely on domain knowledge, manual defect feature construction, and rule application to tackle the defect detection problem, often using rule-based algorithms. Proprietary tools are often used, as it the case for \cite{ben-porath_advanced_1999,ritchison_sem-based_2000,mcgarvey_automated_2009,samir_improvement_2015,cho_classification_2017,wolfe_scanning_2017,egodage_sem_2019}.
There are also some approaches that exploit the periodic nature of wafer patterns \cite{gleason_detection_2002,zontak_kernel-based_2009} and one that uses a reference image to detect defects \cite{lee_accurate_2014}. By leveraging expert knowledge, these methods have achieved high detection and classification accuracies. 

One key advantage of rule-based algorithms is that they can be designed to capture specific patterns and characteristics of defects. For example, rules can be crafted to identify defects based on size, shape, or color. Such rules can be highly effective at identifying defects in certain contexts, where the characteristics of the possible defects are well-defined and known in advance.

However, these approaches have limitations.
A commonly reported issue is that small defects in complex backgrounds or low-contrast, noisy images are missed, as the rules may not be able to distinguish between defects and other features in the image. Another challenge is that the inference time required for these methods can be prohibitively high, making them unsuitable for real-time inspection. 
A more fundamental limitation of these methods is that they rely on expert knowledge and rule-based algorithms, which can be challenging to extend to new situations. For example, if a new type of defect emerges, the expert may need to develop new rules to identify it. This process can be time-consuming, and there is no guarantee that the rules will be effective in all cases. Additionally, rule-based algorithms can be difficult to apply to situations where the defect characteristics are not well-defined.

As manufacturing processes become more complex and the variety of defects increases, the limitations of these methods become more apparent, making it necessary to explore alternative approaches, such as machine learning-based methods.

\subsection{Machine learning based methods}\label{SS: ML}
\begin{figure} []
	\centering
	\includegraphics[width=.9\columnwidth]{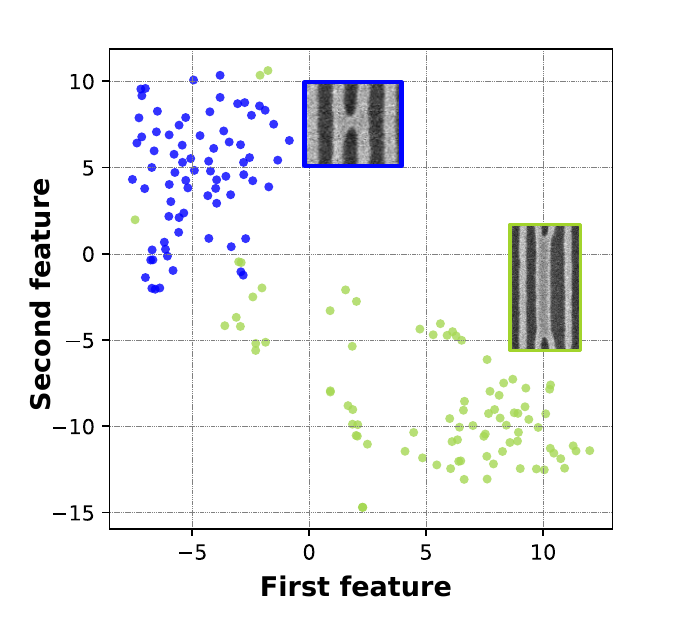}
	\caption{An illustrative example of how k nearest neighbours (kNN) can be used for defect classification. 
	Images of defects were automatically clustered using the \texttt{clustimage} Python package \cite{taskesen_python_2021} and colored according to their corresponding defect class. The features are the result of a Principal Component Analysis \cite{pearson_liii_1901}. Examples of each class are shown in the colored boxes. }
\label{fig:knn}
\end{figure}

The second method category is machine learning based methods, given in Table \ref{tab:ML}. Convolutional neural networks were excluded from this category, as they are quite distinct from other methods in this table. 
In order to learn features and rules, machine learning methods leverage data, which helps to improve their performance \cite{mitchell_machine_1997}.
Such methods include k-nearest neighbours (kNN) \cite{hunt_optimizing_2000}, decision trees \cite{hunt_paradigm_2000}, self-organizing neural networks \cite{kohonen_self-organizing_2001,chang_unsupervised_2005} and Support Vector Machines (SVM's) \cite{cortes_support-vector_1995,graur_new_2015,lee_automated_2020,gomez-sirvent_optimal_2022}.
One of the most popular methods for defect classification is kNN. An example of how this method can be used is given in Fig. \ref{fig:knn}. 
In this example, two features are calculated of over 100 defect instances, each of which is labeled into one of two classes. When a new defect instance needs to be classified, its features are calculated and it is assigned the majority label of the $k$ nearest instances (with $k$ being a natural number). 
Another popular method for classification is an SVM, which is based on the concept of finding a hyperplane that can separate data points of different classes in a high-dimensional space. The support vectors are the data points closest to the hyperplane, and the distance between the hyperplane and the support vectors is maximized in order to achieve better classification performance.

The kNN algorithm works best when there are significant differences between the classes, resulting in easily separable data. 
%Full images are often difficult to use with kNN because they contain high-dimensional data, making it challenging to distinguish between different types of defects. 
Full images are often difficult to use with kNN because they might contain other differences beyond the defects, resulting in data that is not easily distinguishable.
This is why some studies have used cropped images of defects instead, as demonstrated in \cite{hunt_paradigm_2000} and \cite{hunt_optimizing_2000}. 
On the other hand, SVM's are more robust to this issue, which is why they are often used when the data is not easily separable. However, SVM can require more computational resources and is generally less interpretable than kNN.

One of the key factors for kNN and SVM to work correctly is an appropriate distance function.
In high-dimensional problem instances such as images, it can be difficult to find a suitable distance function, which is why dimensionality reduction plays a crucial role. Feature extraction or dimensionality reduction is typically the first step in machine learning-based methods, as it helps to reduce the curse of dimensionality that many methods suffer from~\cite{hughes_mean_1968}. An example is Principal Component Analysis (PCA) \cite{pearson_liii_1901}, which linearly transforms data into a new coordinate system where most of the variation in the data can be captured with fewer dimensions.  

For the task of defect detection, the used methods are more diverse. Examples include using self-organizing neural networks \cite{chang_unsupervised_2005}; automatically aligning a test image with a reference image and taking the pixel-wise difference \cite{chang_similarity_2014}; and extracting local features through histograms of oriented gradients \cite{graur_new_2015}. However, these methods often suffer from poor performance when faced with complex and diverse objects. This is where convolutional neural networks excel, as will be seen in the next section.

Machine learning based methods have shown great potential in defect inspection tasks, especially when a large enough dataset is available. When this is not available, data augmentation techniques can be used to artificially increase the size of the dataset. 
It is also important to have a balanced dataset, meaning that the number of instances in each defect class should be roughly equal. When this is not the case, it can lead to biased model training and poor performance.  Applying data augmentation to the minority class can be used to overcome this issue.
Alternatively, sampling techniques such as SMOTE \cite{chawla_smote_2002} can be used, but they need to be adapted to handle image data well \cite{smith_synthetic_2019}.

Thanks to their ability to learn from data, machine learning based methods are more easily extendable to new patterns or defect types than rule-based algorithms. 
One important consideration is that for supervised machine learning algorithms, an expert is required to label instances before training, which can be time-consuming and expensive. However, it is possible to train unsupervised learning methods without labeled data.

Despite their advantages, machine learning-based methods have some limitations.
These include inference time and bad performance in the case of small datasets or challenging conditions, such as image shifting, low contrast, or high noise levels.
In these cases the performance of machine learning-based methods may be suboptimal, highlighting the need for carefully curated datasets and appropriate preprocessing techniques.

To overcome some of these limitations, researchers are exploring new approaches to machine learning-based defect classification, such as the use of convolutional neural networks and transfer learning. These techniques aim to improve the efficiency and accuracy of machine learning-based defect inspection.

\subsection{Convolutional neural network based methods}\label{SS: CNN}
The third method category is convolutional neural network based methods (CNNs), given in Table \ref{tab:CNN}. These methods have seen a rise in popularity in recent years (see Fig.~\ref{fig:studies over time}) thanks to their performance in object detection \cite{girshick_rich_2014,girshick_fast_2015,ren_faster_2016} and image classification \cite{hirata_ensemble_2020,assiri_stochastic_nodate,liu_convnet_2022}. 
CNNs are a type of deep neural network, composed of layers of interconnected nodes or neurons that perform computations on input data to make predictions or classifications.
 
The key difference between CNNs and `traditional' neural networks like the ones in Table \ref{tab:ML}, is the presence of a large number of layers and a unique type of computation known as a convolution. 
They also require relatively little preprocessing compared to other machine learning methods.
 
There are two crucial building blocks to this type of network, namely convolutional layers and pooling layers. 
Convolutional layers extract feature maps, which contain information about the spatial relationships between the pixels.
These feature maps are then fed into the pooling layers, which apply dimensionality reduction by extracting dominant features~\cite{yang_investigations_2022}.
Formally, each pixel of a feature map can be calculated using
\begin{equation}
FM_{i,j} = (X * C)_{i,j} =  \sum_{h=0}^{h=H-1}\sum_{w=0}^{w=W-1}  X_{i+h,j+w} C_{h,w},
\end{equation}
where $FM_{i,j}$ is the feature map of pixel $(i,j)$ in the resulting matrix, $X$ is the input image, $C$ is the convolution kernel, and $H$ and $W$ correspond to the height and width of the kernel respectively.
Commonly used pooling layers include average pooling and max pooling. 
The latter can be calculated using
\begin{equation}
Y_{i,j} = \max_{(m,n)\in R_{i,j}} X_{m,n},
\end{equation}
where $Y_{i,j}$ is the pooling value of pixel $(i,j)$ in the resulting matrix, over a region $R_{i,j}$ of the input.
Usually one or more fully connected layers are added, which can learn non-linear relationships between these features. This allows the network to recognize complex patterns in the data. Finally, the output layer predicts the probabilities of each class based on the learned features. 

\begin{figure*} [!t]
\centering
\includegraphics[width=.9\textwidth]{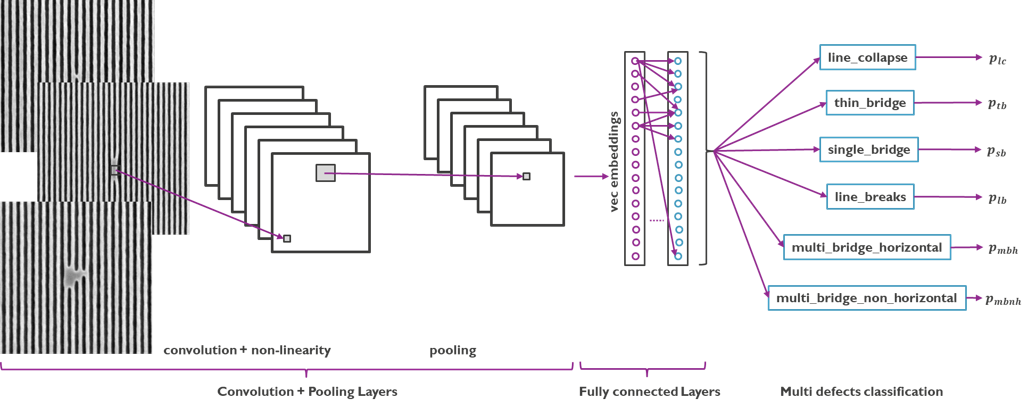}
\caption{An example of a CNN architecture consisting of a convolutional layer, a max pooling layer, a fully connected layer and an output layer. The output contains the estimated probabilities of each defect type.}
\label{fig:CNN}
\end{figure*}

An example of a CNN architecture can be seen in Fig. \ref{fig:CNN}.
This type of architecture can take advantage of the fact that objects are able to be recognized by combining lower-level features, without the need for manual feature extraction. 
In contrast to traditional neural networks, which process each feature in isolation, CNNs take into account the spatial relationships between features.
During the training process, the architecture learns to associate these features with the given defect types. It accomplishes this by iteratively adjusting the weights of the network to minimize a loss function, which measures the difference between the predicted output and the true output. 
Another key aspect of CNNs is the use of weight sharing, which allows the same set of filters to be applied to different parts of the input image. This greatly reduces the number of parameters in the network, making it more efficient and less prone to overfitting.

As CNNs offer a very flexible architecture, they can be optimized for a specific purpose or situation. 
This is highlighted by the fact that every CNN-based study selected in this review used a different architecture or setup. 
%TODO: note that the first block of citations for each method are the included studies that used these methods. The original papers that introduced the methods are cited when explaining the methods.

They ranged from simple ones with only a few convolutional layers, such as in \cite{cheon_convolutional_2019} and \cite{lei_convolutional_2021},
to very complex setups with advanced backbones. Backbones extract features from images and are often quite versatile, in the sense that they can be used in many different setups. For example, they can be used to extract general image features, which are then used by other methods.
Examples of used architectures include Inception \cite{imoto_cnn-based_2019}; ResNet \cite{phua_semiconductor_2020,lin_improvement_2021,kim_adversarial_2021,dey_deep_2022-1,li_automatic_2022}; VGG-16 \cite{fujishiro_minimizing_2021}; R-CNN \cite{he_mask_2018,dey_deep_2022}; U-Net \cite{ofir_automatic_2022}; YOLO \cite{yan_based_2022,dey_towards_2022}; or AlexNet \cite{neumann_defect_2023}. 
The most popular backbone in the included studies is ResNet~\cite{he_deep_2015}, which introduced the concept of residual learning. It adds shortcut connections between layers to allow information to flow more easily through the network. This reduces the vanishing gradient problem \cite{arjovsky_towards_2017} and enables the training of much deeper networks.  
VGG-16 \cite{simonyan_very_2015}, on the other hand, is a convolutional network known for its simplicity and uniformity of design, consisting of 13 convolutional layers followed by 3 fully connected layers. 
R-CNN \cite{girshick_rich_2014}, which stands for Region-based Convolutional Neural Network, is an object detection algorithm that proposes a set of candidate regions in an image and then extracts features from each region using a CNN.
U-Net \cite{ronneberger_u-net_2015} is a CNN-based architecture designed for image segmentation tasks, with a contracting and expansive path for context information and precise localization.
Finally, YOLO \cite{redmon_you_2016}, which stands for You Only Look Once, is a real-time object detection algorithm that is designed to be fast and efficient, with the ability to process images in real-time on a single GPU. There have been multiple improvements to this architecture, both in terms of speed and accuracy. The latest version is currently YOLOv8 \cite{jocher_yolo_2023}.
 %simple: 5, 49,
 %Inception 4, Resnet 44,51, 52,7,61, , VGG-16 6, R-CNN 60, U-net 58,8,
 % Yolo 62, 64, alexNet 67
 %auto encoders: 43,57, 66
 
A different CNN-based approach is the use of autoencoders \cite{dey_unsupervised_2020,fukuda_anomaly_2021,neumann_defect_2023}, which are designed to learn a compressed representation of image data by encoding it into a lower-dimensional space and then reconstructing it back to its original form.
They have been used for anomaly or defect detection by training them on a dataset of normal, non-defective images, and then using the trained model to identify data that deviates significantly from this normal data. The idea is that the autoencoder will learn to compress and reconstruct normal data with high accuracy, and that defective data will be reconstructed less accurately and therefore have a higher reconstruction error.
A threshold is set on the reconstruction error, above which data is considered defective.

The choice of architecture depends on the specific task and dataset, as well as the hardware and computational resources available. 
Training large neural networks requires large amounts of labeled training data, as well as powerful computing resources such as Graphics Processing Units (GPUs) or specialized hardware like Google's Tensor Processing Units (TPUs) \cite{noauthor_train_nodate}. 
However, large training datasets are usually not available for SEM wafer images, at least in terms of real FAB data.  

Two ways of dealing with this problem were commonly seen in the included studies. 
The first is pretraining, where the network can be trained on a large, general dataset such as ImageNet \cite{deng_imagenet_2009} or COCO \cite{lin_microsoft_2015} and later have its parameters fine-tuned on a smaller but specific semiconductor dataset. 
This is commonly referred to as transfer learning, where the goal is to improve the learning of a new task by using information learned from a general task \cite{bozinovski_reminder_2020}. 
A related idea is to use a large dataset with inaccurate labels and later fine-tune the model with reliable labels, as was done in \cite{imoto_cnn-based_2019}.

\begin{figure}[!t]
\centering
%	\subfloat[]{
	\includegraphics[width=3.4in]{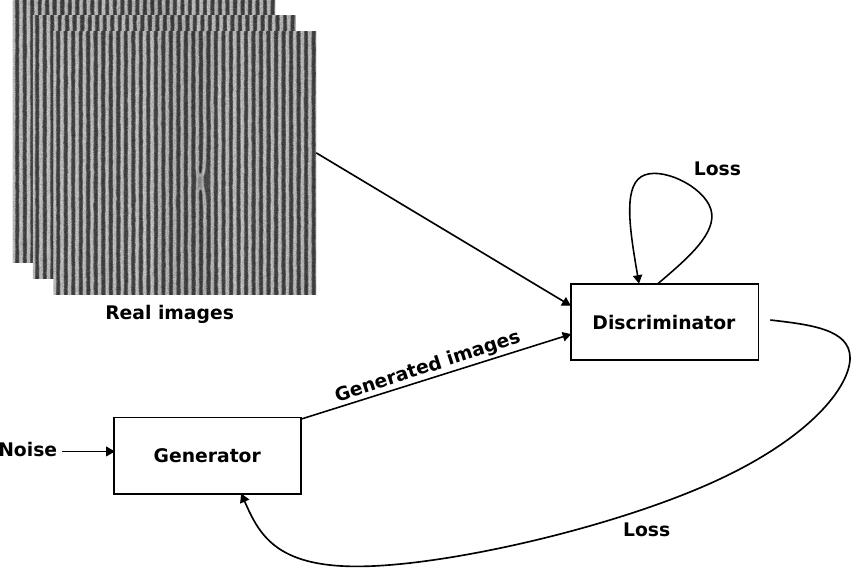}%
	\label{fig: GANarchitecture}
	%	}

%	\subfloat[]{\includegraphics[width=1.6in]{Figures/lc_br_EP_0815.jpg}%
	%		\label{fig: GAN1}}
%	\hfil
%	\subfloat[]{\includegraphics[width=1.6in]{Figures/lc_br_EP_0032.jpg}%
	%		\label{fig: GAN2}}

\caption{An illustration of how generative adversarial networks (GAN's) can be used for data augmentation. A high-level training architecture is shown. }
%		(a) Training architecture (b) A real image. (c) An image generated by a GAN which trained on real images.}
\label{fig: GAN}
\end{figure}

A second way to deal with small training datasets is to use data augmentation. This approach uses slightly modified copies of existing data by shifting, rotating, or adjusting the contrast of those images for example. 
It is also possible to use a Generative Adversarial Network (GAN) \cite{goodfellow_generative_2014} trained on real images to generate synthetic images or to add defects into defect-free images, as was done in \cite{wang_defect_2021}. 
An illustration of how GAN's can be used for data augmentation is given in Figure~\ref{fig: GAN}. A GAN is made up of two neural networks: a generator and a discriminator. The generator creates artificial images, which the discriminator attempts to distinguish from real images. 
The generator improves during training, finally resulting in a model that can generate artificial images which the discriminator is unable to distinguish from real images. These artificial images are then added to the training set, 
aiming to improve the precision and accuracy of the classifier. A major concern is whether or not the artificial data is realistic or has induced digital artifacts, which can negatively influence a trained classifier \cite{zhang_detecting_2019}. Another possible downside is the problem of vanishing gradients \cite{arjovsky_towards_2017}, but this can be remedied by experimenting with different loss functions \cite{goodfellow_generative_2014}.
16 of the 18 CNN-based papers mentioned using some kind of data augmentation or pretraining, so this is a crucial step for training CNNs on (relatively small-scale) semiconductor SEM image datasets.

Overall, CNN-based approaches showed good results thanks to their advanced architectures and efficient use of training data (through e.g. data augmentation and transfer learning). This was the case even for noisy images or when presented with a dataset containing imbalanced classes. 
A limitation of using a CNN for defect inspection is that it requires retraining over time, as new data with new defects are available.
%over time to ensure detection of new defects. 
Additionally, data augmentation does have some limitations for small datasets. The augmented data will carry the biases of the existing dataset, which means a model that trains on this data will often overfit and poorly generalize to new test instances \cite{xu_wemix_2020}.

\subsection{Trends}\label{SS: trends}
As previously mentioned, different methods have seen varying degrees of popularity. A great example is \textbf{convolutional neural networks} (CNNs), as 18 of the 38 publications reviewed were CNN-based, but they were all more recent than 2017. Together with CNNs, the use of \textbf{data augmentation} has also become more popular. 12 of the 18 CNN-based publications use some kind of data augmentation, such as rotation, translation, scaling, contrast and brightness adjustments, or even using GAN's \cite{goodfellow_generative_2014}. 
Data augmentation is beneficial during the training of a CNN, as it can artificially increase the size of the dataset, while keeping the images realistic. This can also be used in other machine learning methods, as is the case in \cite{dey_unsupervised_2020} and \cite{kim_adversarial_2021}.

There is a general trend of an increase in dataset sizes, likely due to two reasons. 
The first is that more data is available now compared to 20 years ago. The second is that larger machine learning algorithms, and specifically CNNs, require larger amounts of training data. Namely, they can learn more complicated patterns with more parameters, but more data is required to fit those parameters.
For instance, the largest dataset in the non-machine learning based studies (Table \ref{tab:nonML}) contains 2077 images \cite{gleason_detection_2002}, while the largest dataset in the CNN-based studies (Table \ref{tab:CNN}) contains 121854 images \cite{imoto_cnn-based_2019}. Thanks to advances in computing power, training large networks on such vast amounts of data has become viable. 

While supervised learning has proved to be very performant, it has the drawback of needing labeled training data. Additionally, when new data or defect types are added, human experts need to label these new images and the models need to be retrained. 
The task of finding anomalies or defects in an unsupervised way is quite challenging however. Only 4 publications in this review proposed unsupervised methods. Namely, \cite{chang_unsupervised_2005} and \cite{chang_using_2005} used an unsupervised neural network in 2005. They proposed methods that can learn to recognize patterns and capture the structure of the input data, which allows them to detect irregularities without requiring labeled examples. 
Later on, \cite{dey_unsupervised_2020} used a variational autoencoder in 2020 and \cite{ofir_automatic_2022} adapted a U-net based CNN to work in an unsupervised setting in 2022. 
Overall, the results were mostly worse than those of supervised methods on the same dataset. For example, the best supervised method tested in \cite{ofir_automatic_2022} achieved an F1 score of $0.87$, while their unsupervised method obtained a score of $0.65$. Therefore, there seems to be a trade-off between accuracy and (lack of) labeling.

Although defect detection and classification methods have become a lot more advanced in recent years, there is no clear trend in their accuracy. This is likely due to the fact that nodes are consistently getting smaller, which makes the defect detection problem harder. Additionally, every study considered in this review uses a different dataset, so their stated accuracies are not directly comparable.

\subsection{Future work}\label{SS: future}
In this section three suggestions towards future work are provided that may improve the capabilities of automated semiconductor defect inspection.

A common limitation of the included studies is that their approach is not easily extensible to new defect types. For example, rule-based methods would require new rules to be added manually and convolutional neural networks would require new, labeled training data. Some studies addressed this issue, for example \cite{cheon_convolutional_2019}, which used kNN on the CNNs feature space in order to classify previously unknown defects. However, even in this case the CNN needs retraining over time to ensure detection of new defects, which is computationally expensive. A possible solution to this problem could be to use unsupervised machine learning methods, which do not require labeled training data. Unsupervised anomaly detection in particular has seen lots of innovation in recent years, see e.g. \cite{defard_padim_2020,cohen_sub-image_2021,gudovskiy_cflow-ad_2021,yu_fastflow_2021}.
The 4 publications included in this review that use unsupervised methods have not yet been able to match the performance of supervised methods, but
more advanced unsupervised architectures or learning methods might be able to. Such methods have already been successfully applied to other fields \cite{hamilton_unsupervised_2022} and they have the potential to avoid a large amount of human labeling and the biases that come with it.
Furthermore, using semi-supervised \cite{sohn_simple_2020} or few-shot learning methods \cite{kang_few-shot_2019,sun_fsce_2021} can facilitate the progression towards fully unsupervised methods by reducing the reliance on labeled training data. This would also help to avoid the limitations associated with data augmentation. 
%This is also linked to making the training step of CNNs more efficient. For instance, 12 of the 13 CNN-based methods used some kind of pretraining or data augmentation. 

Another common limitation is inference time, as detecting defects can be a bottleneck in the manufacturing process \cite{ben-porath_advanced_1999,yan_based_2022}. Analyzing the trade-off between inference time and accuracy could therefore be an interesting avenue of further research. This could involve studying the performance of smaller models or using an ensemble of methods that can detect defects faster than a singular, larger model. An example of such a study is \cite{li_automatic_2022}, where a decision tree of 8 smaller CNNs is shown to be faster than a single large CNN.
Furthermore, compacting a large neural network in order to reduce inference time, but still achieving satisfactory accuracy might be useful for deploying such methods in a production environment \cite{kuo_advanced_2015,zhang_compacting_2021}.

% Fix limitations of the method categories by e.g. combining
Although every publication included in this review reported great accuracy results, these are not directly comparable as they all used different datasets. The development of a public semiconductor defect dataset would therefore be extremely useful. This would allow new defect detection methods to be properly benchmarked, foster innovation and contribute to the development of accurate and reliable defect inspection methods. An example of an industrial anomaly detection dataset is MVTec AD \cite{bergmann_mvtec_2019}, which has been used in over 146 papers \cite{noauthor_papers_nodate} and has helped establish the effectiveness of different approaches. Note that there is an annotated set of SEM images \cite{aversa_first_2018}, but it is not semiconductor-specific and does not contain defect information.

% CNNs: supervised in two ways: labeled data + specific architecture for the problem at hand.
% -> work towards a general architecture?
% Make training more efficient, as 12 of the 13 CNN-based methods used some kind of pretraining or data augmentation.

%Model centric approach: 

% Table: category vs application (rows: nonML,ML,CNN , columns: detection, classification, localization/segmentation)

\section{Conclusion}
As the semiconductor industry moves towards smaller nodes, the challenge of detecting and classifying semiconductor defects using SEM images is becoming a lot harder.
These defects may affect electrical performance and reliability, therefore identifying and addressing them is crucial. 
Manual inspection methods can be time-consuming and prone to human error, so using automated detection methods can be more efficient and accurate. 
Additionally, the use of automated methods can reduce the time and resources required for defect detection, making it a valuable inspection tool for semiconductor industry.
A large amount of research has been dedicated to propose solutions to this problem, particularly in recent years. 

This systematic review selected 38 studies from IEEE Xplore and SPIE, published between 1995 and 2023. Their applications, methodologies, datasets, results, limitations and future work were summarized and their general findings were discussed. Three main method categories were identified, namely non-machine learning-based, convolutional neural network-based and machine learning-based methods (excluding CNNs).
%The most important applications were illustrated and trends in methodology were observed.
For each of these, an overview was given on their use cases, main advantages and limitations. Key trends were identified, for example the recent popularity of CNNs or the usage of data augmentation strategies. 
Finally, promising directions for future work were suggested. Namely, unsupervised machine learning methods should be investigated more thoroughly as they have the potential to avoid the tedious human labeling requirement and the biases that come with it. The trade-off between inference time and accuracy should be analyzed, and a public semiconductor defect dataset should be developed.

\bibliographystyle{IEEEtran}
\bibliography{bibliography}

\begin{thebibliography}{100}
\providecommand{\url}[1]{#1}
\csname url@samestyle\endcsname
\providecommand{\newblock}{\relax}
\providecommand{\bibinfo}[2]{#2}
\providecommand{\BIBentrySTDinterwordspacing}{\spaceskip=0pt\relax}
\providecommand{\BIBentryALTinterwordstretchfactor}{4}
\providecommand{\BIBentryALTinterwordspacing}{\spaceskip=\fontdimen2\font plus
\BIBentryALTinterwordstretchfactor\fontdimen3\font minus
  \fontdimen4\font\relax}
\providecommand{\BIBforeignlanguage}[2]{{%
\expandafter\ifx\csname l@#1\endcsname\relax
\typeout{** WARNING: IEEEtran.bst: No hyphenation pattern has been}%
\typeout{** loaded for the language `#1'. Using the pattern for}%
\typeout{** the default language instead.}%
\else
\language=\csname l@#1\endcsname
\fi
#2}}
\providecommand{\BIBdecl}{\relax}
\BIBdecl

\bibitem{moore_cramming_1965}
G.~E. Moore, ``\BIBforeignlanguage{en}{Cramming more components onto integrated
  circuits},'' \emph{\BIBforeignlanguage{en}{Electronics}}, vol.~38, no.~8,
  1965.

\bibitem{dey_deep_2022-3}
\BIBentryALTinterwordspacing
B.~Dey, E.~Dehaerne, K.~Khalil, S.~Halder, P.~Leray, and M.~A. Bayoumi, ``Deep
  {Learning} based {Defect} classification and detection in {SEM} images: {A}
  {Mask} {R}-{CNN} approach,'' Nov. 2022, arXiv:2211.02185 [cs]. [Online].
  Available: \url{http://arxiv.org/abs/2211.02185}
\BIBentrySTDinterwordspacing

\bibitem{dey_deep_2022-2}
B.~Dey, D.~Goswamif, S.~Haldera, K.~Khalilb, P.~Leraya, and M.~Bayoumi,
  \emph{Deep {Learning}-{Based} {Defect} {Classification} and {Detection} in
  {SEM} {Images}}, Jun. 2022.

\bibitem{dehaerne_optimizing_2023}
\BIBentryALTinterwordspacing
E.~Dehaerne, B.~Dey, S.~Halder, and S.~De~Gendt, ``Optimizing {YOLOv7} for
  {Semiconductor} {Defect} {Detection},'' Feb. 2023. [Online]. Available:
  \url{http://arxiv.org/abs/2302.09565}
\BIBentrySTDinterwordspacing

\bibitem{breaux_automatic_1995}
L.~Breaux and B.~Singh, ``Automatic defect classification system for patterned
  semiconductor wafers,'' in \emph{Proceedings of {International} {Symposium}
  on {Semiconductor} {Manufacturing}}, Sep. 1995, pp. 68--73.

\bibitem{imoto_cnn-based_2019}
K.~Imoto, T.~Nakai, T.~Ike, K.~Haruki, and Y.~Sato, ``A {CNN}-{Based}
  {Transfer} {Learning} {Method} for {Defect} {Classification} in
  {Semiconductor} {Manufacturing},'' \emph{IEEE Transactions on Semiconductor
  Manufacturing}, vol.~32, no.~4, pp. 455--459, Nov. 2019.

\bibitem{cheon_convolutional_2019}
S.~Cheon, H.~Lee, C.~O. Kim, and S.~H. Lee, ``Convolutional {Neural} {Network}
  for {Wafer} {Surface} {Defect} {Classification} and the {Detection} of
  {Unknown} {Defect} {Class},'' \emph{IEEE Transactions on Semiconductor
  Manufacturing}, vol.~32, no.~2, pp. 163--170, May 2019.

\bibitem{fujishiro_minimizing_2021}
A.~Fujishiro, Y.~Nagamura, T.~Usami, and M.~Inoue, ``Minimizing {Convolutional}
  {Neural} {Network} {Training} {Data} {With} {Proper} {Data} {Augmentation}
  for {Inline} {Defect} {Classification},'' \emph{IEEE Transactions on
  Semiconductor Manufacturing}, vol.~34, no.~3, pp. 333--339, Aug. 2021.

\bibitem{dey_deep_2022-1}
B.~Dey, D.~Goswami, S.~Halder, K.~Khalil, P.~Leray, and M.~A. Bayoumi, ``Deep
  learning-based defect classification and detection in {SEM} images,'' in
  \emph{Metrology, {Inspection}, and {Process} {Control} {XXXVI}}, vol.
  PC12053.\hskip 1em plus 0.5em minus 0.4em\relax SPIE, Jun. 2022, p.
  PC120530Y.

\bibitem{ofir_automatic_2022}
N.~Ofir, R.~Yacobi, O.~Granoviter, B.~Levant, and O.~Shtalrid, ``Automatic
  {Defect} {Segmentation} by {Unsupervised} {Anomaly} {Learning},'' in
  \emph{2022 {IEEE} {International} {Conference} on {Image} {Processing}
  ({ICIP})}, Oct. 2022, pp. 306--310.

\bibitem{batool_systematic_2021}
U.~Batool, M.~I. Shapiai, M.~Tahir, Z.~H. Ismail, N.~J. Zakaria, and
  A.~Elfakharany, ``A {Systematic} {Review} of {Deep} {Learning} for {Silicon}
  {Wafer} {Defect} {Recognition},'' \emph{IEEE Access}, vol.~9, pp.
  116\,572--116\,593, 2021.

\bibitem{theodosiou_review_2023}
\BIBentryALTinterwordspacing
T.~Theodosiou, A.~Rapti, K.~Papageorgiou, T.~Tziolas, E.~Papageorgiou,
  N.~Dimitriou, G.~Margetis, and D.~Tzovaras, ``\BIBforeignlanguage{en}{A
  {Review} {Study} on {ML}-based {Methods} for {Defect}-{Pattern} {Recognition}
  in {Wafer} {Maps}},'' \emph{\BIBforeignlanguage{en}{Procedia Computer
  Science}}, vol. 217, pp. 570--583, Jan. 2023. [Online]. Available:
  \url{https://www.sciencedirect.com/science/article/pii/S1877050922023316}
\BIBentrySTDinterwordspacing

\bibitem{ma_review_2023}
\BIBentryALTinterwordspacing
J.~Ma, T.~Zhang, C.~Yang, Y.~Cao, L.~Xie, H.~Tian, and X.~Li,
  ``\BIBforeignlanguage{en}{Review of {Wafer} {Surface} {Defect} {Detection}
  {Methods}},'' \emph{\BIBforeignlanguage{en}{Electronics}}, vol.~12, no.~8, p.
  1787, Jan. 2023. [Online]. Available:
  \url{https://www.mdpi.com/2079-9292/12/8/1787}
\BIBentrySTDinterwordspacing

\bibitem{zheng_recent_2021}
\BIBentryALTinterwordspacing
X.~Zheng, S.~Zheng, Y.~Kong, and J.~Chen, ``\BIBforeignlanguage{en}{Recent
  advances in surface defect inspection of industrial products using deep
  learning techniques},'' \emph{\BIBforeignlanguage{en}{The International
  Journal of Advanced Manufacturing Technology}}, vol. 113, no.~1, pp. 35--58,
  Mar. 2021. [Online]. Available:
  \url{https://doi.org/10.1007/s00170-021-06592-8}
\BIBentrySTDinterwordspacing

\bibitem{lopez_de_la_rosa_review_2021}
\BIBentryALTinterwordspacing
F.~López de~la Rosa, R.~Sánchez-Reolid, J.~L. Gómez-Sirvent, R.~Morales, and
  A.~Fernández-Caballero, ``\BIBforeignlanguage{en}{A {Review} on {Machine}
  and {Deep} {Learning} for {Semiconductor} {Defect} {Classification} in
  {Scanning} {Electron} {Microscope} {Images}},''
  \emph{\BIBforeignlanguage{en}{Applied Sciences}}, vol.~11, no.~20, p. 9508,
  Jan. 2021. [Online]. Available:
  \url{https://www.mdpi.com/2076-3417/11/20/9508}
\BIBentrySTDinterwordspacing

\bibitem{page_prisma_2021}
\BIBentryALTinterwordspacing
M.~J. Page, J.~E. McKenzie, P.~M. Bossuyt, I.~Boutron, T.~C. Hoffmann, C.~D.
  Mulrow, L.~Shamseer, J.~M. Tetzlaff, E.~A. Akl, S.~E. Brennan, R.~Chou,
  J.~Glanville, J.~M. Grimshaw, A.~Hróbjartsson, M.~M. Lalu, T.~Li, E.~W.
  Loder, E.~Mayo-Wilson, S.~McDonald, L.~A. McGuinness, L.~A. Stewart,
  J.~Thomas, A.~C. Tricco, V.~A. Welch, P.~Whiting, and D.~Moher,
  ``\BIBforeignlanguage{en}{The {PRISMA} 2020 statement: an updated guideline
  for reporting systematic reviews},'' \emph{\BIBforeignlanguage{en}{BMJ}},
  vol. 372, p. n71, Mar. 2021. [Online]. Available:
  \url{https://www.bmj.com/content/372/bmj.n71}
\BIBentrySTDinterwordspacing

\bibitem{noauthor_ieee_nodate}
\BIBentryALTinterwordspacing
``{IEEE} {Xplore}: {Advanced} {Search}.'' [Online]. Available:
  \url{https://ieeexplore.ieee.org/search/advanced/command}
\BIBentrySTDinterwordspacing

\bibitem{noauthor_spie_nodate}
\BIBentryALTinterwordspacing
``{SPIE} {Digital} {Library}.'' [Online]. Available:
  \url{https://www.spiedigitallibrary.org/?SSO=1}
\BIBentrySTDinterwordspacing

\bibitem{noauthor_crossref_nodate}
\BIBentryALTinterwordspacing
``\BIBforeignlanguage{en}{Crossref}.'' [Online]. Available:
  \url{https://www.crossref.org/}
\BIBentrySTDinterwordspacing

\bibitem{ben-porath_advanced_1999}
A.~Ben-Porath, T.~Hayes, and A.~Skumanich, ``Advanced process development and
  control based on a fully automated {SEM} with {ADC},'' in \emph{10th {Annual}
  {IEEE}/{SEMI}. {Advanced} {Semiconductor} {Manufacturing} {Conference} and
  {Workshop}. {ASMC} 99 {Proceedings} ({Cat}. {No}.{99CH36295})}, Sep. 1999,
  pp. 275--280.

\bibitem{ritchison_sem-based_2000}
\BIBentryALTinterwordspacing
J.~W. Ritchison, A.~Ben-Porath, and E.~Malocsay, ``{SEM}-based {ADC} evaluation
  and integration in an advanced process fab,'' in \emph{Metrology,
  {Inspection}, and {Process} {Control} for {Microlithography} {XIV}}, vol.
  3998.\hskip 1em plus 0.5em minus 0.4em\relax SPIE, Jun. 2000, pp. 258--268.
  [Online]. Available:
  \url{https://www.spiedigitallibrary.org/conference-proceedings-of-spie/3998/0000/SEM-based-ADC-evaluation-and-integration-in-an-advanced-process/10.1117/12.386480.full}
\BIBentrySTDinterwordspacing

\bibitem{gleason_detection_2002}
S.~S. Gleason, R.~K. Ferrell, T.~P. Karnowski, and K.~W.~T. Jr, ``Detection of
  semiconductor defects using a novel fractal encoding algorithm,'' in
  \emph{Design, {Process} {Integration}, and {Characterization} for
  {Microelectronics}}, vol. 4692.\hskip 1em plus 0.5em minus 0.4em\relax SPIE,
  Jul. 2002, pp. 61--71.

\bibitem{mcgarvey_automated_2009}
\BIBentryALTinterwordspacing
S.~McGarvey and M.~Kanezawa, ``Automated defect review of the wafer bevel with
  a defect review scanning electron microscope,'' in \emph{Metrology,
  {Inspection}, and {Process} {Control} for {Microlithography} {XXIII}}, vol.
  7272.\hskip 1em plus 0.5em minus 0.4em\relax SPIE, Mar. 2009, pp. 1077--1088.
  [Online]. Available:
  \url{https://www.spiedigitallibrary.org/conference-proceedings-of-spie/7272/72723N/Automated-defect-review-of-the-wafer-bevel-with-a-defect/10.1117/12.814235.full}
\BIBentrySTDinterwordspacing

\bibitem{zontak_kernel-based_2009}
M.~Zontak and I.~Cohen, ``Kernel-based detection of defects on semiconductor
  wafers,'' in \emph{2009 {IEEE} {International} {Workshop} on {Machine}
  {Learning} for {Signal} {Processing}}, Sep. 2009, pp. 1--6.

\bibitem{lee_accurate_2014}
Y.~Lee and J.~Lee, ``Accurate {Automatic} {Defect} {Detection} {Method} {Using}
  {Quadtree} {Decomposition} on {SEM} {Images},'' \emph{IEEE Transactions on
  Semiconductor Manufacturing}, vol.~27, no.~2, pp. 223--231, May 2014.

\bibitem{samir_improvement_2015}
B.~Samir, M.~Pereira, S.~Paninjath, C.-U. Jeon, D.-H. Chung, G.-S. Yoon, and
  H.-Y. Jung, ``Improvement in accuracy of defect size measurement by automatic
  defect classification,'' in \emph{Photomask {Technology} 2015}, vol.
  9635.\hskip 1em plus 0.5em minus 0.4em\relax SPIE, Oct. 2015, pp. 477--483.

\bibitem{cho_classification_2017}
\BIBentryALTinterwordspacing
W.~Cho, D.~Price, P.~A. Morgan, D.~Rost, M.~Satake, and V.~L. Tolani,
  ``Classification and printability of {EUV} mask defects from {SEM} images,''
  in \emph{International {Conference} on {Extreme} {Ultraviolet} {Lithography}
  2017}, vol. 10450.\hskip 1em plus 0.5em minus 0.4em\relax SPIE, Oct. 2017,
  pp. 28--35. [Online]. Available:
  \url{https://www.spiedigitallibrary.org/conference-proceedings-of-spie/10450/1045006/Classification-and-printability-of-EUV-mask-defects-from-SEM-images/10.1117/12.2280837.full}
\BIBentrySTDinterwordspacing

\bibitem{wolfe_scanning_2017}
S.~Wolfe and S.~McGarvey, ``Scanning electron microscope automatic defect
  classification of process induced defects,'' in \emph{Metrology,
  {Inspection}, and {Process} {Control} for {Microlithography} {XXXI}}, vol.
  10145.\hskip 1em plus 0.5em minus 0.4em\relax SPIE, Mar. 2017, pp. 622--635.

\bibitem{egodage_sem_2019}
\BIBentryALTinterwordspacing
K.~Egodage, F.~Tu, H.~Schneider, C.~F. Hermanns, G.~Kersteen, B.~Szafranek, and
  K.~Schulz, ``{SEM} {AutoAnalysis} for reduced turnaround time and to ensure
  repair quality of {EUV} photomasks,'' in \emph{International {Conference} on
  {Extreme} {Ultraviolet} {Lithography} 2019}, vol. 11147.\hskip 1em plus 0.5em
  minus 0.4em\relax SPIE, Sep. 2019, pp. 156--165. [Online]. Available:
  \url{https://www.spiedigitallibrary.org/conference-proceedings-of-spie/11147/111471G/SEM-AutoAnalysis-for-reduced-turnaround-time-and-to-ensure-repair/10.1117/12.2538474.full}
\BIBentrySTDinterwordspacing

\bibitem{hunt_paradigm_2000}
M.~A. Hunt, J.~S.~G. Jr, J.~A. Mullens, R.~K. Ferrell, B.~R. Whitus, and
  A.~Ben-Porath, ``Paradigm for selecting the optimum classifier in
  semiconductor automatic defect classification applications,'' in
  \emph{Metrology, {Inspection}, and {Process} {Control} for {Microlithography}
  {XIV}}, vol. 3998.\hskip 1em plus 0.5em minus 0.4em\relax SPIE, Jun. 2000,
  pp. 269--276.

\bibitem{hunt_optimizing_2000}
M.~Hunt, T.~Karnowski, C.~Kiest, and L.~Villalobos, ``Optimizing automatic
  defect classification feature and classifier performance for post-fab yield
  analysis,'' in \emph{2000 {IEEE}/{SEMI} {Advanced} {Semiconductor}
  {Manufacturing} {Conference} and {Workshop}. {ASMC} 2000 ({Cat}.
  {No}.{00CH37072})}, Sep. 2000, pp. 116--123.

\bibitem{chang_unsupervised_2005}
C.-Y. Chang, J.-W. Chang, and M.~d. Jeng, ``An {Unsupervised}
  {Self}-{Organizing} {Neural} {Network} for {Automatic} {Semiconductor}
  {Wafer} {Defect} {Inspection},'' in \emph{Proceedings of the 2005 {IEEE}
  {International} {Conference} on {Robotics} and {Automation}}, Apr. 2005, pp.
  3000--3005.

\bibitem{kohonen_self-organizing_2001}
\BIBentryALTinterwordspacing
T.~Kohonen, \emph{Self-{Organizing} {Maps}}, ser. Springer {Series} in
  {Information} {Sciences}, T.~S. Huang, T.~Kohonen, and M.~R. Schroeder,
  Eds.\hskip 1em plus 0.5em minus 0.4em\relax Berlin, Heidelberg: Springer,
  2001, vol.~30. [Online]. Available:
  \url{http://link.springer.com/10.1007/978-3-642-56927-2}
\BIBentrySTDinterwordspacing

\bibitem{chang_using_2005}
C.-Y. Chang, S.-Y. Lin, and M.~Jeng, ``Using a two-layer competitive {Hopfield}
  neural network for semiconductor wafer defect detection,'' in \emph{{IEEE}
  {International} {Conference} on {Automation} {Science} and {Engineering},
  2005.}, Aug. 2005, pp. 301--306.

\bibitem{hopfield_neural_1982}
\BIBentryALTinterwordspacing
J.~J. Hopfield, ``Neural networks and physical systems with emergent collective
  computational abilities.'' \emph{Proceedings of the National Academy of
  Sciences of the United States of America}, vol.~79, no.~8, pp. 2554--2558,
  Apr. 1982. [Online]. Available:
  \url{https://www.ncbi.nlm.nih.gov/pmc/articles/PMC346238/}
\BIBentrySTDinterwordspacing

\bibitem{chang_similarity_2014}
J.-Y. Chang and W.-L. Chang, ``Similarity based {ART} 1 model for automatic die
  defect detection and classification,'' in \emph{2014 {Joint} 7th
  {International} {Conference} on {Soft} {Computing} and {Intelligent}
  {Systems} ({SCIS}) and 15th {International} {Symposium} on {Advanced}
  {Intelligent} {Systems} ({ISIS})}, Dec. 2014, pp. 883--887.

\bibitem{noauthor_mean_2008}
\BIBentryALTinterwordspacing
``\BIBforeignlanguage{en}{Mean {Absolute} {Deviation}},'' in
  \emph{\BIBforeignlanguage{en}{The {Concise} {Encyclopedia} of
  {Statistics}}}.\hskip 1em plus 0.5em minus 0.4em\relax New York, NY:
  Springer, 2008, pp. 336--337. [Online]. Available:
  \url{https://doi.org/10.1007/978-0-387-32833-1_250}
\BIBentrySTDinterwordspacing

\bibitem{carpenter_adaptive_2010}
\BIBentryALTinterwordspacing
G.~A. Carpenter and S.~Grossberg, ``\BIBforeignlanguage{en}{Adaptive
  {Resonance} {Theory}},'' in \emph{\BIBforeignlanguage{en}{Encyclopedia of
  {Machine} {Learning}}}, C.~Sammut and G.~I. Webb, Eds.\hskip 1em plus 0.5em
  minus 0.4em\relax Boston, MA: Springer US, 2010, pp. 22--35. [Online].
  Available: \url{https://doi.org/10.1007/978-0-387-30164-8_11}
\BIBentrySTDinterwordspacing

\bibitem{graur_new_2015}
\BIBentryALTinterwordspacing
I.~Graur, D.~Vengertsev, A.~Raghunathan, I.~Stobert, and J.~Rankin, ``New
  method of detection and classification of yield-impacting {EUV} mask
  defects,'' in \emph{Photomask {Technology} 2015}, vol. 9635.\hskip 1em plus
  0.5em minus 0.4em\relax SPIE, Oct. 2015, pp. 114--120. [Online]. Available:
  \url{https://www.spiedigitallibrary.org/conference-proceedings-of-spie/9635/96350M/New-method-of-detection-and-classification-of-yield-impacting-EUV/10.1117/12.2197871.full}
\BIBentrySTDinterwordspacing

\bibitem{rana_deep_2018}
\BIBentryALTinterwordspacing
N.~Rana and C.~Chien, ``Deep machine learning based {Image} classification in
  hard disk drive manufacturing,'' in \emph{Metrology, {Inspection}, and
  {Process} {Control} for {Microlithography} {XXXII}}, vol. 10585.\hskip 1em
  plus 0.5em minus 0.4em\relax SPIE, Apr. 2018, pp. 217--224. [Online].
  Available:
  \url{https://www.spiedigitallibrary.org/conference-proceedings-of-spie/10585/105850Y/Deep-machine-learning-based-Image-classification-in-hard-disk-drive/10.1117/12.2300952.full}
\BIBentrySTDinterwordspacing

\bibitem{lee_automated_2020}
P.-H. Lee, Z.~Wang, C.~Teh, Y.-S. Hsiao, and W.~Fang, ``Automated semiconductor
  wafer defect classification dealing with imbalanced data,'' in
  \emph{Metrology, {Inspection}, and {Process} {Control} for {Microlithography}
  {XXXIV}}, vol. 11325.\hskip 1em plus 0.5em minus 0.4em\relax SPIE, Mar. 2020,
  pp. 381--387.

\bibitem{chawla_smote_2002}
\BIBentryALTinterwordspacing
N.~V. Chawla, K.~W. Bowyer, L.~O. Hall, and W.~P. Kegelmeyer,
  ``\BIBforeignlanguage{en}{{SMOTE}: {Synthetic} {Minority} {Over}-sampling
  {Technique}},'' \emph{\BIBforeignlanguage{en}{Journal of Artificial
  Intelligence Research}}, vol.~16, pp. 321--357, Jun. 2002. [Online].
  Available: \url{https://www.jair.org/index.php/jair/article/view/10302}
\BIBentrySTDinterwordspacing

\bibitem{ho_random_1995}
T.~K. Ho, ``Random decision forests,'' in \emph{Proceedings of 3rd
  {International} {Conference} on {Document} {Analysis} and {Recognition}},
  vol.~1, Aug. 1995, pp. 278--282 vol.1.

\bibitem{freund_desicion-theoretic_1995}
Y.~Freund and R.~E. Schapire, ``\BIBforeignlanguage{en}{A desicion-theoretic
  generalization of on-line learning and an application to boosting},'' in
  \emph{\BIBforeignlanguage{en}{Computational {Learning} {Theory}}}, ser.
  Lecture {Notes} in {Computer} {Science}, P.~Vitányi, Ed.\hskip 1em plus
  0.5em minus 0.4em\relax Berlin, Heidelberg: Springer, 1995, pp. 23--37.

\bibitem{chen_xgboost_2016}
\BIBentryALTinterwordspacing
T.~Chen and C.~Guestrin, ``{XGBoost}: {A} {Scalable} {Tree} {Boosting}
  {System},'' in \emph{Proceedings of the 22nd {ACM} {SIGKDD} {International}
  {Conference} on {Knowledge} {Discovery} and {Data} {Mining}}, Aug. 2016, pp.
  785--794. [Online]. Available: \url{http://arxiv.org/abs/1603.02754}
\BIBentrySTDinterwordspacing

\bibitem{cortes_support-vector_1995}
\BIBentryALTinterwordspacing
C.~Cortes and V.~Vapnik, ``\BIBforeignlanguage{en}{Support-vector networks},''
  \emph{\BIBforeignlanguage{en}{Machine Learning}}, vol.~20, no.~3, pp.
  273--297, Sep. 1995. [Online]. Available:
  \url{https://doi.org/10.1007/BF00994018}
\BIBentrySTDinterwordspacing

\bibitem{gomez-sirvent_optimal_2022}
J.~L. Gómez-Sirvent, F.~L. de~la Rosa, R.~Sánchez-Reolid,
  A.~Fernández-Caballero, and R.~Morales, ``Optimal {Feature} {Selection} for
  {Defect} {Classification} in {Semiconductor} {Wafers},'' \emph{IEEE
  Transactions on Semiconductor Manufacturing}, vol.~35, no.~2, pp. 324--331,
  May 2022.

\bibitem{ducote_microlens_2018}
\BIBentryALTinterwordspacing
J.~Ducoté, A.~Lakcher, L.~Bidault, A.-R. Philipot, A.~Ostrovsky, E.~Mortini,
  and B.~Le-Gratiet, ``Microlens under melt in-line monitoring based on
  application of neural network automatic defect classification,'' in
  \emph{34th {European} {Mask} and {Lithography} {Conference}}, vol.
  10775.\hskip 1em plus 0.5em minus 0.4em\relax SPIE, Sep. 2018, pp. 112--115.
  [Online]. Available:
  \url{https://www.spiedigitallibrary.org/conference-proceedings-of-spie/10775/107750S/Microlens-under-melt-in-line-monitoring-based-on-application-of/10.1117/12.2326397.full}
\BIBentrySTDinterwordspacing

\bibitem{szegedy_rethinking_2015}
\BIBentryALTinterwordspacing
C.~Szegedy, V.~Vanhoucke, S.~Ioffe, J.~Shlens, and Z.~Wojna, ``Rethinking the
  {Inception} {Architecture} for {Computer} {Vision},'' Dec. 2015. [Online].
  Available: \url{http://arxiv.org/abs/1512.00567}
\BIBentrySTDinterwordspacing

\bibitem{dey_unsupervised_2020}
B.~Dey, D.~Cerbu, K.~Khalil, S.~Halder, P.~Leray, S.~Das, Y.~Sherazi, M.~A.
  Bayoumi, and R.~H. Kim, ``Unsupervised machine learning based {CD}-{SEM}
  image segregator for {OPC} and process window estimation,'' in
  \emph{Design-{Process}-{Technology} {Co}-optimization for {Manufacturability}
  {XIV}}, vol. 11328.\hskip 1em plus 0.5em minus 0.4em\relax SPIE, Mar. 2020,
  pp. 317--327.

\bibitem{phua_semiconductor_2020}
C.~Phua and L.~B. Theng, ``Semiconductor {Wafer} {Surface}: {Automatic}
  {Defect} {Classification} with {Deep} {CNN},'' in \emph{2020 {IEEE} {REGION}
  10 {CONFERENCE} ({TENCON})}, Nov. 2020, pp. 714--719.

\bibitem{he_deep_2015}
\BIBentryALTinterwordspacing
K.~He, X.~Zhang, S.~Ren, and J.~Sun, ``Deep {Residual} {Learning} for {Image}
  {Recognition},'' Dec. 2015. [Online]. Available:
  \url{http://arxiv.org/abs/1512.03385}
\BIBentrySTDinterwordspacing

\bibitem{deng_imagenet_2009}
J.~Deng, W.~Dong, R.~Socher, L.-J. Li, K.~Li, and L.~Fei-Fei, ``{ImageNet}: {A}
  large-scale hierarchical image database,'' in \emph{2009 {IEEE} {Conference}
  on {Computer} {Vision} and {Pattern} {Recognition}}, Jun. 2009, pp. 248--255.

\bibitem{liu_ssd_2016}
W.~Liu, D.~Anguelov, D.~Erhan, C.~Szegedy, S.~Reed, C.-Y. Fu, and A.~C. Berg,
  ``\BIBforeignlanguage{en}{{SSD}: {Single} {Shot} {MultiBox} {Detector}},'' in
  \emph{\BIBforeignlanguage{en}{Computer {Vision} – {ECCV} 2016}}, ser.
  Lecture {Notes} in {Computer} {Science}, B.~Leibe, J.~Matas, N.~Sebe, and
  M.~Welling, Eds.\hskip 1em plus 0.5em minus 0.4em\relax Cham: Springer
  International Publishing, 2016, pp. 21--37.

\bibitem{simonyan_very_2015}
\BIBentryALTinterwordspacing
K.~Simonyan and A.~Zisserman, ``Very {Deep} {Convolutional} {Networks} for
  {Large}-{Scale} {Image} {Recognition},'' Apr. 2015. [Online]. Available:
  \url{http://arxiv.org/abs/1409.1556}
\BIBentrySTDinterwordspacing

\bibitem{lei_convolutional_2021}
H.~Lei, C.~Teh, Z.~Wang, G.~Fu, L.~Pu, and W.~Fang, ``Convolutional {Neural}
  {Network} ({CNN}) {Based} {Automated} {Defect} {Classification} ({ADC}) with
  {Imbalanced} {Data},'' in \emph{2021 {China} {Semiconductor} {Technology}
  {International} {Conference} ({CSTIC})}, Mar. 2021, pp. 1--3.

\bibitem{goodfellow_generative_2014}
\BIBentryALTinterwordspacing
I.~Goodfellow, J.~Pouget-Abadie, M.~Mirza, B.~Xu, D.~Warde-Farley, S.~Ozair,
  A.~Courville, and Y.~Bengio, ``Generative {Adversarial} {Nets},'' in
  \emph{Advances in {Neural} {Information} {Processing} {Systems}},
  vol.~27.\hskip 1em plus 0.5em minus 0.4em\relax Curran Associates, Inc.,
  2014. [Online]. Available:
  \url{https://proceedings.neurips.cc/paper/2014/hash/5ca3e9b122f61f8f06494c97b1afccf3-Abstract.html}
\BIBentrySTDinterwordspacing

\bibitem{lin_improvement_2021}
B.-S. Lin, J.-S. Cheng, H.-C. Liao, L.-W. Yang, T.~Yang, and K.-C. Chen,
  ``Improvement of {Multi}-{Lines} {Bridge} {Defect} {Classification} by
  {Hierarchical} {Architecture} in {Artificial} {Intelligence} {Automatic}
  {Defect} {Classification},'' \emph{IEEE Transactions on Semiconductor
  Manufacturing}, vol.~34, no.~3, pp. 346--351, Aug. 2021.

\bibitem{kim_adversarial_2021}
J.~Kim, Y.~Nam, M.-C. Kang, K.~Kim, J.~Hong, S.~Lee, and D.-N. Kim,
  ``Adversarial {Defect} {Detection} in {Semiconductor} {Manufacturing}
  {Process},'' \emph{IEEE Transactions on Semiconductor Manufacturing},
  vol.~34, no.~3, pp. 365--371, Aug. 2021.

\bibitem{mirza_conditional_2014}
\BIBentryALTinterwordspacing
M.~Mirza and S.~Osindero, ``Conditional {Generative} {Adversarial} {Nets},''
  Nov. 2014. [Online]. Available: \url{http://arxiv.org/abs/1411.1784}
\BIBentrySTDinterwordspacing

\bibitem{lin_focal_2018}
\BIBentryALTinterwordspacing
T.-Y. Lin, P.~Goyal, R.~Girshick, K.~He, and P.~Dollár, ``Focal {Loss} for
  {Dense} {Object} {Detection},'' Feb. 2018. [Online]. Available:
  \url{http://arxiv.org/abs/1708.02002}
\BIBentrySTDinterwordspacing

\bibitem{makhzani_pixelgan_2017}
\BIBentryALTinterwordspacing
A.~Makhzani and B.~Frey, ``{PixelGAN} {Autoencoders},'' Jun. 2017. [Online].
  Available: \url{http://arxiv.org/abs/1706.00531}
\BIBentrySTDinterwordspacing

\bibitem{wang_defect_2021}
Z.~Wang, L.~Yu, and L.~Pu, ``Defect simulation in {SEM} images using generative
  adversarial networks,'' in \emph{Metrology, {Inspection}, and {Process}
  {Control} for {Semiconductor} {Manufacturing} {XXXV}}, vol. 11611.\hskip 1em
  plus 0.5em minus 0.4em\relax SPIE, Feb. 2021, pp. 113--119.

\bibitem{fukuda_anomaly_2021}
H.~Fukuda and T.~Kondo, ``Anomaly detection in random circuit patterns using
  autoencoder,'' \emph{Journal of Micro/Nanopatterning, Materials, and
  Metrology}, vol.~20, no.~4, p. 044001, Nov. 2021.

\bibitem{ronneberger_u-net_2015}
\BIBentryALTinterwordspacing
O.~Ronneberger, P.~Fischer, and T.~Brox, ``U-{Net}: {Convolutional} {Networks}
  for {Biomedical} {Image} {Segmentation},'' May 2015. [Online]. Available:
  \url{http://arxiv.org/abs/1505.04597}
\BIBentrySTDinterwordspacing

\bibitem{dey_deep_2022}
B.~Dey, E.~Dehaerne, S.~Halder, P.~Leray, and M.~A. Bayoumi, ``Deep learning
  based defect classification and detection in {SEM} images: a mask {R}-{CNN}
  approach,'' in \emph{Metrology, {Inspection}, and {Process} {Control}
  {XXXVI}}, vol. PC12053.\hskip 1em plus 0.5em minus 0.4em\relax SPIE, Jul.
  2022, p. PC120530K.

\bibitem{he_mask_2018}
\BIBentryALTinterwordspacing
K.~He, G.~Gkioxari, P.~Dollár, and R.~Girshick, ``Mask {R}-{CNN},'' Jan. 2018.
  [Online]. Available: \url{http://arxiv.org/abs/1703.06870}
\BIBentrySTDinterwordspacing

\bibitem{li_automatic_2022}
Z.~Li, Z.~Wang, and W.~Shi, ``Automatic {Wafer} {Defect} {Classification}
  {Based} on {Decision} {Tree} of {Deep} {Neural} {Network},'' in \emph{2022
  33rd {Annual} {SEMI} {Advanced} {Semiconductor} {Manufacturing} {Conference}
  ({ASMC})}, May 2022, pp. 1--6.

\bibitem{yan_based_2022}
S.~Yan, S.~Ding, S.~Wang, C.~Luo, L.~Li, J.~Ai, Q.~Shen, Q.~Xia, Z.~Li,
  Q.~Cheng, S.~Li, H.~Dai, and X.~Hu, ``Based on {Deep} {Learning} {CD}-{SEM}
  {Image} {Defect} {Detection} {System},'' in \emph{2022 {China}
  {Semiconductor} {Technology} {International} {Conference} ({CSTIC})}, Jun.
  2022, pp. 1--4.

\bibitem{redmon_yolo9000_2016}
\BIBentryALTinterwordspacing
J.~Redmon and A.~Farhadi, ``{YOLO9000}: {Better}, {Faster}, {Stronger},'' Dec.
  2016. [Online]. Available: \url{http://arxiv.org/abs/1612.08242}
\BIBentrySTDinterwordspacing

\bibitem{dey_towards_2022}
B.~Dey, E.~Dehaerne, and S.~Halder, ``\BIBforeignlanguage{en}{Towards improving
  challenging stochastic defect detection in {SEM} images based on improved
  {YOLOv5}},'' in \emph{\BIBforeignlanguage{en}{Photomask {Technology} 2022}},
  B.~S. Kasprowicz and T.~Liang, Eds.\hskip 1em plus 0.5em minus 0.4em\relax
  Monterey, United States: SPIE, Dec. 2022, p.~11.

\bibitem{jocher_ultralyticsyolov5_2022}
\BIBentryALTinterwordspacing
G.~Jocher, A.~Chaurasia, A.~Stoken, J.~Borovec, NanoCode012, Y.~Kwon,
  K.~Michael, TaoXie, J.~Fang, imyhxy, Lorna, Z.~Yifu, C.~Wong, A.~V,
  D.~Montes, Z.~Wang, C.~Fati, J.~Nadar, Laughing, UnglvKitDe, V.~Sonck,
  tkianai, yxNONG, P.~Skalski, A.~Hogan, D.~Nair, M.~Strobel, and M.~Jain,
  ``ultralytics/yolov5: v7.0 - {YOLOv5} {SOTA} {Realtime} {Instance}
  {Segmentation},'' Nov. 2022. [Online]. Available:
  \url{https://zenodo.org/record/7347926}
\BIBentrySTDinterwordspacing

\bibitem{neumann_defect_2023}
J.~T. Neumann, A.~Srikantha, P.~Hüthwohl, K.~Lee, J.~William~B., T.~Korb,
  E.~Foca, T.~Garbowski, D.~Boecker, S.~Das, and S.~Halder,
  ``\BIBforeignlanguage{en}{Defect detection and classification on imec {iN5}
  node {BEoL} test vehicle with multibeam scanning electron microscope},''
  \emph{\BIBforeignlanguage{en}{Journal of Micro/Nanopatterning, Materials, and
  Metrology}}, vol.~22, no.~02, Mar. 2023.

\bibitem{krizhevsky_imagenet_2012}
\BIBentryALTinterwordspacing
A.~Krizhevsky, I.~Sutskever, and G.~E. Hinton, ``{ImageNet} {Classification}
  with {Deep} {Convolutional} {Neural} {Networks},'' in \emph{Advances in
  {Neural} {Information} {Processing} {Systems}}, vol.~25.\hskip 1em plus 0.5em
  minus 0.4em\relax Curran Associates, Inc., 2012. [Online]. Available:
  \url{https://papers.nips.cc/paper_files/paper/2012/hash/c399862d3b9d6b76c8436e924a68c45b-Abstract.html}
\BIBentrySTDinterwordspacing

\bibitem{pedregosa_scikit-learn_2011}
F.~Pedregosa, G.~Varoquaux, A.~Gramfort, V.~Michel, B.~Thirion, O.~Grisel,
  M.~Blondel, P.~Prettenhofer, R.~Weiss, V.~Dubourg, J.~Vanderplas, A.~Passos,
  D.~Cournapeau, M.~Brucher, M.~Perrot, and E.~Duchesnay, ``Scikit-learn:
  {Average} {Precision} {Score} metric,'' \emph{Journal of Machine Learning
  Research}, vol.~12, pp. 2825--2830, 2011.

\bibitem{anwar_average_2022}
\BIBentryALTinterwordspacing
A.~Anwar, ``\BIBforeignlanguage{en}{Average {Precision} in {Object} {Detection}
  \& {Localization} {Algorithms} and how to calculate it},'' May 2022.
  [Online]. Available:
  \url{https://towardsdatascience.com/what-is-average-precision-in-object-detection-localization-algorithms-and-how-to-calculate-it-3f330efe697b}
\BIBentrySTDinterwordspacing

\bibitem{esposito_analysis_2020}
T.~A. Esposito, S.-H. Jen, Q.~Xie, D.~Acharya, J.~Lee, and F.~Levitov,
  ``Analysis of {Systematic} {Weak} {Point} {Structures} using {Design} {Based}
  {Automatic} {Defect} {Classification} and {Defect} {Review} {SEM}
  {Platform},'' in \emph{2020 31st {Annual} {SEMI} {Advanced} {Semiconductor}
  {Manufacturing} {Conference} ({ASMC})}, Aug. 2020, pp. 1--4.

\bibitem{shah_design_2018}
J.~Shah, A.~Jain, F.~Levitov, S.~Yasharzade, J.~G. Sheridan, V.~Nguyen, and
  H.~Nguyen, ``Design based automatic defect classification at advanced
  technology nodes: {DI}: {Defect} inspection and reduction,'' in \emph{2018
  29th {Annual} {SEMI} {Advanced} {Semiconductor} {Manufacturing} {Conference}
  ({ASMC})}, Apr. 2018, pp. 270--275.

\bibitem{oleary_deep_2020}
J.~O’Leary, K.~Sawlani, and A.~Mesbah, ``Deep {Learning} for {Classification}
  of the {Chemical} {Composition} of {Particle} {Defects} on {Semiconductor}
  {Wafers},'' \emph{IEEE Transactions on Semiconductor Manufacturing}, vol.~33,
  no.~1, pp. 72--85, Feb. 2020.

\bibitem{arena_exploiting_2021}
S.~Arena, Y.~Bodrov, M.~Carletti, N.~Gentner, M.~Maggipinto, Y.~Yang, A.~Beghi,
  A.~Kyek, and G.~A. Susto, ``Exploiting {2D} {Coordinates} as {Bayesian}
  {Priors} for {Deep} {Learning} {Defect} {Classification} of {SEM} {Images},''
  \emph{IEEE Transactions on Semiconductor Manufacturing}, vol.~34, no.~3, pp.
  436--439, Aug. 2021.

\bibitem{mitchell_machine_1997}
T.~M. Mitchell, \emph{\BIBforeignlanguage{en}{Machine {Learning}}}, ser.
  {McGraw}-{Hill} series in computer science.\hskip 1em plus 0.5em minus
  0.4em\relax New York: McGraw-Hill, 1997.

\bibitem{hughes_mean_1968}
G.~Hughes, ``On the mean accuracy of statistical pattern recognizers,''
  \emph{IEEE Transactions on Information Theory}, vol.~14, no.~1, pp. 55--63,
  Jan. 1968.

\bibitem{pearson_liii_1901}
\BIBentryALTinterwordspacing
K.~Pearson, ``{LIII}. {On} lines and planes of closest fit to systems of points
  in space,'' \emph{The London, Edinburgh, and Dublin Philosophical Magazine
  and Journal of Science}, vol.~2, no.~11, pp. 559--572, Nov. 1901. [Online].
  Available: \url{https://doi.org/10.1080/14786440109462720}
\BIBentrySTDinterwordspacing

\bibitem{smith_synthetic_2019}
\BIBentryALTinterwordspacing
T.~Smith and J.~White, ``\BIBforeignlanguage{en}{Synthetic {Minority}
  {Reconstruction} {Technique}},'' 2019. [Online]. Available:
  \url{https://github.com/tgsmith61591/smrt}
\BIBentrySTDinterwordspacing

\bibitem{taskesen_python_2021}
\BIBentryALTinterwordspacing
E.~Taskesen, ``Python package clustimage is for unsupervised clustering of
  images.'' Nov. 2021. [Online]. Available:
  \url{https://erdogant.github.io/clustimage}
\BIBentrySTDinterwordspacing

\bibitem{girshick_rich_2014}
R.~Girshick, J.~Donahue, T.~Darrell, and J.~Malik, ``Rich {Feature}
  {Hierarchies} for {Accurate} {Object} {Detection} and {Semantic}
  {Segmentation},'' in \emph{2014 {IEEE} {Conference} on {Computer} {Vision}
  and {Pattern} {Recognition}}, Jun. 2014, pp. 580--587.

\bibitem{girshick_fast_2015}
\BIBentryALTinterwordspacing
R.~Girshick, ``Fast {R}-{CNN},'' Sep. 2015. [Online]. Available:
  \url{http://arxiv.org/abs/1504.08083}
\BIBentrySTDinterwordspacing

\bibitem{ren_faster_2016}
\BIBentryALTinterwordspacing
S.~Ren, K.~He, R.~Girshick, and J.~Sun, ``Faster {R}-{CNN}: {Towards}
  {Real}-{Time} {Object} {Detection} with {Region} {Proposal} {Networks},''
  Jan. 2016. [Online]. Available: \url{http://arxiv.org/abs/1506.01497}
\BIBentrySTDinterwordspacing

\bibitem{hirata_ensemble_2020}
\BIBentryALTinterwordspacing
D.~Hirata and N.~Takahashi, ``Ensemble learning in {CNN} augmented with fully
  connected subnetworks,'' Mar. 2020. [Online]. Available:
  \url{http://arxiv.org/abs/2003.08562}
\BIBentrySTDinterwordspacing

\bibitem{assiri_stochastic_nodate}
Y.~S. Assiri, ``\BIBforeignlanguage{en}{Stochastic {Optimization} of {Plain}
  {Convolutional} {Neural} {Networks} with {Simple} methods}.''

\bibitem{liu_convnet_2022}
\BIBentryALTinterwordspacing
Z.~Liu, H.~Mao, C.-Y. Wu, C.~Feichtenhofer, T.~Darrell, and S.~Xie, ``A
  {ConvNet} for the 2020s,'' Mar. 2022. [Online]. Available:
  \url{http://arxiv.org/abs/2201.03545}
\BIBentrySTDinterwordspacing

\bibitem{yang_investigations_2022}
\BIBentryALTinterwordspacing
A.~Yang, M.~Liu, J.~Wang, Q.~He, N.~Zhang, and J.~Jia,
  ``\BIBforeignlanguage{en}{Investigations of {Object} {Detectors} with {Deep}
  {Learning} {Methods}: {A} {Review}},'' in \emph{\BIBforeignlanguage{en}{2022
  5th {International} {Conference} on {Pattern} {Recognition} and {Artificial}
  {Intelligence} ({PRAI})}}.\hskip 1em plus 0.5em minus 0.4em\relax Chengdu,
  China: IEEE, Aug. 2022, pp. 351--358. [Online]. Available:
  \url{https://ieeexplore.ieee.org/document/9904222/}
\BIBentrySTDinterwordspacing

\bibitem{arjovsky_towards_2017}
\BIBentryALTinterwordspacing
M.~Arjovsky and L.~Bottou, ``Towards {Principled} {Methods} for {Training}
  {Generative} {Adversarial} {Networks},'' Jan. 2017. [Online]. Available:
  \url{http://arxiv.org/abs/1701.04862}
\BIBentrySTDinterwordspacing

\bibitem{redmon_you_2016}
\BIBentryALTinterwordspacing
J.~Redmon, S.~Divvala, R.~Girshick, and A.~Farhadi, ``You {Only} {Look} {Once}:
  {Unified}, {Real}-{Time} {Object} {Detection},'' May 2016. [Online].
  Available: \url{http://arxiv.org/abs/1506.02640}
\BIBentrySTDinterwordspacing

\bibitem{jocher_yolo_2023}
\BIBentryALTinterwordspacing
G.~Jocher, A.~Chaurasia, and J.~Qiu, ``{YOLO} by {Ultralytics},'' Jan. 2023.
  [Online]. Available: \url{https://github.com/ultralytics/ultralytics}
\BIBentrySTDinterwordspacing

\bibitem{noauthor_train_nodate}
\BIBentryALTinterwordspacing
``\BIBforeignlanguage{en}{Train and run machine learning models faster
  {\textbar} {Cloud} {TPU}}.'' [Online]. Available:
  \url{https://cloud.google.com/tpu}
\BIBentrySTDinterwordspacing

\bibitem{lin_microsoft_2015}
\BIBentryALTinterwordspacing
T.-Y. Lin, M.~Maire, S.~Belongie, L.~Bourdev, R.~Girshick, J.~Hays, P.~Perona,
  D.~Ramanan, C.~L. Zitnick, and P.~Dollár, ``Microsoft {COCO}: {Common}
  {Objects} in {Context},'' Feb. 2015. [Online]. Available:
  \url{http://arxiv.org/abs/1405.0312}
\BIBentrySTDinterwordspacing

\bibitem{bozinovski_reminder_2020}
\BIBentryALTinterwordspacing
S.~Bozinovski, ``\BIBforeignlanguage{en}{Reminder of the {First} {Paper} on
  {Transfer} {Learning} in {Neural} {Networks}, 1976},''
  \emph{\BIBforeignlanguage{en}{Informatica}}, vol.~44, no.~3, Sep. 2020.
  [Online]. Available:
  \url{http://www.informatica.si/index.php/informatica/article/view/2828}
\BIBentrySTDinterwordspacing

\bibitem{zhang_detecting_2019}
\BIBentryALTinterwordspacing
X.~Zhang, S.~Karaman, and S.-F. Chang, ``Detecting and {Simulating} {Artifacts}
  in {GAN} {Fake} {Images},'' Oct. 2019. [Online]. Available:
  \url{http://arxiv.org/abs/1907.06515}
\BIBentrySTDinterwordspacing

\bibitem{xu_wemix_2020}
\BIBentryALTinterwordspacing
Y.~Xu, A.~Noy, M.~Lin, Q.~Qian, H.~Li, and R.~Jin, ``{WeMix}: {How} to {Better}
  {Utilize} {Data} {Augmentation},'' Oct. 2020. [Online]. Available:
  \url{http://arxiv.org/abs/2010.01267}
\BIBentrySTDinterwordspacing

\bibitem{defard_padim_2020}
\BIBentryALTinterwordspacing
T.~Defard, A.~Setkov, A.~Loesch, and R.~Audigier, ``{PaDiM}: a {Patch}
  {Distribution} {Modeling} {Framework} for {Anomaly} {Detection} and
  {Localization},'' Nov. 2020. [Online]. Available:
  \url{http://arxiv.org/abs/2011.08785}
\BIBentrySTDinterwordspacing

\bibitem{cohen_sub-image_2021}
\BIBentryALTinterwordspacing
N.~Cohen and Y.~Hoshen, ``Sub-{Image} {Anomaly} {Detection} with {Deep}
  {Pyramid} {Correspondences},'' Feb. 2021. [Online]. Available:
  \url{http://arxiv.org/abs/2005.02357}
\BIBentrySTDinterwordspacing

\bibitem{gudovskiy_cflow-ad_2021}
\BIBentryALTinterwordspacing
D.~Gudovskiy, S.~Ishizaka, and K.~Kozuka, ``{CFLOW}-{AD}: {Real}-{Time}
  {Unsupervised} {Anomaly} {Detection} with {Localization} via {Conditional}
  {Normalizing} {Flows},'' Jul. 2021. [Online]. Available:
  \url{http://arxiv.org/abs/2107.12571}
\BIBentrySTDinterwordspacing

\bibitem{yu_fastflow_2021}
\BIBentryALTinterwordspacing
J.~Yu, Y.~Zheng, X.~Wang, W.~Li, Y.~Wu, R.~Zhao, and L.~Wu, ``{FastFlow}:
  {Unsupervised} {Anomaly} {Detection} and {Localization} via {2D}
  {Normalizing} {Flows},'' Nov. 2021. [Online]. Available:
  \url{http://arxiv.org/abs/2111.07677}
\BIBentrySTDinterwordspacing

\bibitem{hamilton_unsupervised_2022}
M.~Hamilton, Z.~Zhang, B.~Hariharan, N.~Snavely, and W.~T. Freeman,
  ``\BIBforeignlanguage{en}{Unsupervised semantic segmentation by distilling
  feature correspondences},'' 2022.

\bibitem{sohn_simple_2020}
\BIBentryALTinterwordspacing
K.~Sohn, Z.~Zhang, C.-L. Li, H.~Zhang, C.-Y. Lee, and T.~Pfister, ``A {Simple}
  {Semi}-{Supervised} {Learning} {Framework} for {Object} {Detection},'' Dec.
  2020. [Online]. Available: \url{http://arxiv.org/abs/2005.04757}
\BIBentrySTDinterwordspacing

\bibitem{kang_few-shot_2019}
\BIBentryALTinterwordspacing
B.~Kang, Z.~Liu, X.~Wang, F.~Yu, J.~Feng, and T.~Darrell, ``Few-shot {Object}
  {Detection} via {Feature} {Reweighting},'' Oct. 2019. [Online]. Available:
  \url{http://arxiv.org/abs/1812.01866}
\BIBentrySTDinterwordspacing

\bibitem{sun_fsce_2021}
\BIBentryALTinterwordspacing
B.~Sun, B.~Li, S.~Cai, Y.~Yuan, and C.~Zhang, ``{FSCE}: {Few}-{Shot} {Object}
  {Detection} via {Contrastive} {Proposal} {Encoding},'' Mar. 2021. [Online].
  Available: \url{http://arxiv.org/abs/2103.05950}
\BIBentrySTDinterwordspacing

\bibitem{kuo_advanced_2015}
Y.~Kuo, Y.-L. Fang, H.-C. Liao, L.-W. Yang, T.~Yang, and K.-C. Chen, ``Advanced
  high accuracy scanning electron microscopy review methodology by virtual
  defect — {Yiting} {Kuo},'' in \emph{2015 {Joint} e-{Manufacturing} and
  {Design} {Collaboration} {Symposium} ({eMDC}) \& 2015 {International}
  {Symposium} on {Semiconductor} {Manufacturing} ({ISSM})}, Sep. 2015, pp.
  1--4.

\bibitem{zhang_compacting_2021}
\BIBentryALTinterwordspacing
K.~Zhang, H.~Ying, H.-N. Dai, L.~Li, Y.~Peng, K.~Guo, and H.~Yu, ``Compacting
  {Deep} {Neural} {Networks} for {Internet} of {Things}: {Methods} and
  {Applications},'' \emph{IEEE Internet of Things Journal}, vol.~8, no.~15, pp.
  11\,935--11\,959, Aug. 2021. [Online]. Available:
  \url{http://arxiv.org/abs/2103.11083}
\BIBentrySTDinterwordspacing

\bibitem{bergmann_mvtec_2019}
\BIBentryALTinterwordspacing
P.~Bergmann, M.~Fauser, D.~Sattlegger, and C.~Steger,
  ``\BIBforeignlanguage{en}{{MVTec} {AD} — {A} {Comprehensive} {Real}-{World}
  {Dataset} for {Unsupervised} {Anomaly} {Detection}},'' in
  \emph{\BIBforeignlanguage{en}{2019 {IEEE}/{CVF} {Conference} on {Computer}
  {Vision} and {Pattern} {Recognition} ({CVPR})}}.\hskip 1em plus 0.5em minus
  0.4em\relax Long Beach, CA, USA: IEEE, Jun. 2019, pp. 9584--9592. [Online].
  Available: \url{https://ieeexplore.ieee.org/document/8954181/}
\BIBentrySTDinterwordspacing

\bibitem{noauthor_papers_nodate}
\BIBentryALTinterwordspacing
``\BIBforeignlanguage{en}{Papers with {Code} - {MVTecAD} {Dataset}}.''
  [Online]. Available: \url{https://paperswithcode.com/dataset/mvtecad}
\BIBentrySTDinterwordspacing

\bibitem{aversa_first_2018}
\BIBentryALTinterwordspacing
R.~Aversa, M.~H. Modarres, S.~Cozzini, R.~Ciancio, and A.~Chiusole,
  ``\BIBforeignlanguage{en}{The first annotated set of scanning electron
  microscopy images for nanoscience},''
  \emph{\BIBforeignlanguage{en}{Scientific Data}}, vol.~5, no.~1, p. 180172,
  Aug. 2018. [Online]. Available:
  \url{https://www.nature.com/articles/sdata2018172}
\BIBentrySTDinterwordspacing

\end{thebibliography}

\end{document}